%% file: bare_jrnl_compsoc.tex
\pdfoutput=1

\documentclass[10pt,journal,compsoc]{IEEEtran}
\ifCLASSOPTIONcompsoc
  \usepackage[nocompress]{cite}
\else
  \usepackage{cite}
\fi
\usepackage{graphicx}
\usepackage{enumerate}
\usepackage{bm}
\usepackage{amsmath}
\usepackage{subfig}
\usepackage{multirow}
\usepackage{ifpdf}
\ifCLASSINFOpdf
\else
\fi
\begin{document}
\title{Isomorphic mesh generation from point clouds with multilayer perceptrons}

\author{Shoko~Miyauchi,~\IEEEmembership{Member,~IEEE,}
        Ken'ichi~Morooka,~\IEEEmembership{Member,~IEEE,}
        and~Ryo~Kurazume,~\IEEEmembership{Member,~IEEE}% <-this % stops a space
\IEEEcompsocitemizethanks{\IEEEcompsocthanksitem S. Miyauchi and R. Kurazume are with the Graduate School of Information Science and Electrical Engineering, Kyushu University, Fukuoka,
Japan, 8190395.\protect\\
E-mail: miyauchi@ait.kyushu-u.ac.jp
\IEEEcompsocthanksitem K. Morooka is with the Faculty of Engineering, Okayama University, Okayama, Japan, 7008530.}% <-this % stops an unwanted space
\thanks{}}


\markboth{Journal of \LaTeX\ Class Files,~Vol.~14, No.~8, August~2015}%
{Shell \MakeLowercase{\textit{et al.}}: Bare Demo of IEEEtran.cls for Computer Society Journals}



\IEEEtitleabstractindextext{%
\begin{abstract}
%Three-dimensional (3D) object models obtained by a mobile device facilitate the application of deep neural networks (DNNs) to analyse 3D shapes. 
%However, the number of points in each model is different from each other, and the connection between the points is unknown.
We propose a new neural network, called isomorphic mesh generator (iMG), which generates isomorphic meshes from point clouds containing noise and missing parts.
%as one solution for dealing with such object models with DNNs while considering the connecting relationship.
Isomorphic meshes of arbitrary objects have a unified mesh structure even though the objects belong to different classes.
%Isomorphic meshes have an advantage of extracting 3D shape features while considering their vertex-to-vertex connectivity without increasing memory usage and calculation time.
This unified representation enables surface models to be handled by DNNs. 
%Since the spatial relationships between points of the objects can be utilized when training DNNs, it is possible to extract shape features of 3D object models while taking into account those surface shapes.
%Compared with the point cloud,
%since 
%mesh models including isomorphic meshes
%contain the spatial relationships between points of the objects,
%the features obtained by using the spatial relationships
%are powerful tools for DNN-based applications using 3D object models.
Moreover, the unified mesh structure of isomorphic meshes enables the same process to be applied to all isomorphic meshes; although
in the case of general mesh models,
we need to consider 
the processes depending on their mesh structures. 
Therefore, the use of isomorphic meshes
leads to efficient memory usage and calculation time compared with general mesh models.
As iMG is a data-free method, 
preparing any point clouds as training data in advance is unnecessary, except a point cloud of the target object used as the input data of iMG.
Additionally, iMG outputs an isomorphic mesh obtained by mapping a reference mesh to a given input point cloud. 
To estimate the mapping function stably, we introduce a step-by-step mapping strategy.
This strategy achieves a flexible deformation while maintaining the structure of the reference mesh.
From simulation and experiments using a mobile phone, we confirmed that iMG can generate isomorphic meshes of given objects reliably even when the input point cloud includes noise and missing parts.
\end{abstract}

}

\maketitle

\IEEEdisplaynontitleabstractindextext

% \fi
\IEEEpeerreviewmaketitle

\input{introduction}

\input{relatedwork}

\input{method}

\input{experiment}

\input{discussion}

\input{conclusion}
\ifCLASSOPTIONcompsoc
  \section*{Acknowledgments}
\else
  \section*{Acknowledgment}
\fi

This work was supported by JST, ACT-X Grant Number JPMJAX190T and CREST Grant Number JPMJCR20F3, Japan.
Special thanks to Kyo Itaya of Okayama University for his support towards the experiment.

\ifCLASSOPTIONcaptionsoff
  \newpage
\fi



\bibliographystyle{IEEEtran}
\bibliography{related_work}
\end{document}

%% file: introduction.tex
\section{Introduction}
\label{intro}
In recent years, mobile devices, such as iPhones, have been equipped with depth sensors that capture three-dimensional (3D) point clouds of objects and natural scenes.
The use of these mobile devices with a depth sensor increases the opportunity to deal with 3D object models generated from the 3D point cloud obtained using the depth sensor.
Furthermore, 3D object models facilitate the application of deep neural networks (DNNs) to classify, segment, and reconstruct 3D shapes and surfaces \cite{liu2021pointguard, wen2021airborne, liu2021fine, mirbauer2021survey, azcona2020interpretation, lakhili2019deformable, george2022deep, huang20213d, qiu2021dense, wang2021learning, xu2020weakly, han2020occuseg, yi2019gspn, wang2019graph, yu2019partnet, meng2019vv, jiang2018pointsift, su2018splatnet, hanocka2019meshcnn, chibane2020implicit, genova2020local, chabra2020deep, chen2020bsp, gao2019sdm, chen2019learning}.

In general, the input data of a DNN needs to be represented by a fixed-size vector or array, which fixes the architecture of the input layer of the DNN.
Therefore, when applying DNNs to 3D object models, the models must be described by a vector or array with a specific dimension.
Additionally, the vector or array elements of the 3D object model should have a well-defined permutation to correspond to and compare with other 3D object models.

Here, the number of points in each model is different even in the case of object models in the same class. 
As DNNs that can handle such a point cloud directly, there are 
PointNet-based methods \cite{qi2017pointnet, qi2017pointnet++, xie2021generative, luo2021knn, sheshappanavar2020novel} that satisfy permutation invariance.
Here, a point cloud consists of the positions of points sampled from the surface of an object. 
On the contrary, the point cloud contains no information about the spatial relationship between the points.
If the relationships between points such as a geodesic can be utilized when training DNNs, for example, the shape features of the 3D object models can be extracted while considering their surface shapes.
Such features have the potential to improve the accuracy of DNN-based applications using 3D objects.
One method to enable DNNs to handle 3D object models while considering such relationships is to convert those point clouds to mesh models in advance.
However, when the structures of surface models are different from each other, DNNs cannot handle them directly as inputs.

One solution for handling 3D object models using DNNs while considering the connecting relationship is to convert the point cloud into a surface mesh with a unified structure.
This indicates that the surface meshes of all objects have the same number of vertices and mesh structure. 
One advantage of the surface meshes, called 'isomorphic meshes',
is that they represent all such meshes with a specific dimensional vector.  
This unified representation enables the DNNs to handle these surface models.
Moreover, the unified mesh structure of isomorphic meshes allows the same process to be applied to all isomorphic meshes, although
in the case of general mesh models,
we need to consider 
the processes depending on their corresponding mesh structures. 
Therefore, the use of isomorphic meshes
leads to efficient memory usage and calculation time, unlike the use of general mesh models.

Various methods have been developed to generate a surface mesh from the point cloud \cite{kazhdan2006poisson, williams2021neural, jiang2020local, ummenhofer2021adaptive, zhang2021learning, azinovic2021neural, atzmon2020sal, mi2020ssrnet}.
However, few previous mesh generation methods considered generating an isomorphic mesh with a fixed number of vertices.
Moreover, the point clouds obtained by the sensor sometimes contain noise and missing parts. 
Therefore, the method of generating isomorphic meshes needs to recover the object shape while filling the missing parts and removing noise.

To solve these problems, we propose a DNN-based method to generate isomorphic meshes from object point clouds.
The proposed method, called isomorphic mesh generator (iMG), is based on a deep geometric prior (DGP) \cite{williams2019deep}, which divides the point cloud into multiple local regions and outputs a dense and smooth point cloud of each local region. 
The DGP is a data-free method that needs no other data except a given input data for constructing the DGP.
Moreover, the DGP is robust to noise and missing parts.
Accordingly, the DGP is applicable to point clouds with noise and missing parts.
However, the purpose of the DGP is to increase the density of the point cloud.
Unlike the DGP, the proposed iMG generates an isomorphic mesh from the point cloud of an object. 
Specifically, when the point cloud of an object and a reference mesh are provided, the iMG outputs the deformed reference mesh, which recovers the shape of the point clouds.
Using this same reference mesh, we can describe the input point clouds of different objects with the mesh structure of the reference mesh.

In the model generation,  iMG deforms the reference mesh to fit the point cloud distribution of the target object.
When the shape of the target object is largely different from that of the reference mesh, the reference mesh is deformed drastically to fit the shape of the target object.
Such a drastic deformation sometimes results in generating an incomplete mesh that includes many self-intersections.
One solution for the self-intersection during deformation is to prepare different reference meshes according to the shape of the point cloud \cite{hanocka2020point2mesh}.
However, this solution requires additional pre-processing to determine the reference mesh with a suitable shape for the input point cloud.
Moreover, if the reference meshes have different mesh structures,
these meshes 
need to be converted into a unified mesh structure to obtain isomorphic meshes.

\begin{figure*}[tb]
\centering
		\includegraphics[clip, width=1.6\columnwidth]{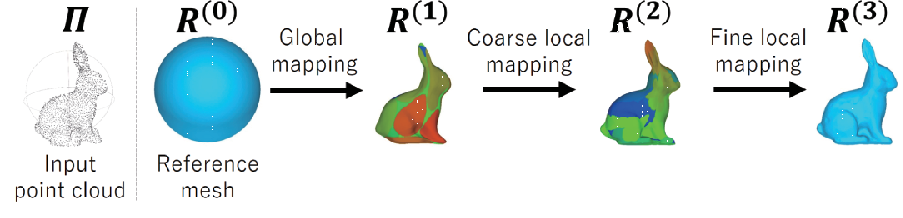}
	\caption{Overview of our iMG algorithm. Different colours in $\bm R^{(1)}$ and $\bm R^{(2)}$ are assigned to each local region}
	\label{fig:method}
\end{figure*}

To solve the problem, iMG performs three-step mapping, i.e., global mapping, coarse local mapping, and fine local mapping (see Fig. \ref{fig:method}).
First, given a point cloud of a target object, global mapping approximately recovers the entire shape of the target object by deforming the reference mesh.
Second, after dividing the point cloud into several local regions, coarse local mapping is performed to recover the approximate shape of each local region independently.
Third, fine local mapping is performed to represent the detailed shape of the cloud to improve the shape-recovery accuracy.
Here, the shape-recovery accuracy indicates the degree of shape matching between
the target object and reference mesh obtained by mapping.
The three-step mapping achieves fine and stable shape recovery of the target object when using a common reference mesh whose shape is different from that of the target object.

%% file: relatedwork.tex
\section{Related work}
Previous studies on reconstructing
surface meshes from point clouds
can be classified into two categories according to the use of neural networks: classical methods that do not use neural networks and neural network-based methods.

\subsection{Classical methods without using neural networks}
A simple method for reconstructing surface meshes includes a Voronoi-based algorithm \cite{amenta1998new} and ball-pivoting algorithm \cite{bernardini1999ball} that generates a mesh by connecting two points with an edge.
However, the accuracy of the mesh generated by such simple methods is sensitive to noise and missing parts because of the direct use of the given point clouds.

Another mesh-generation method approximates the local surface region of the point cloud using parametric
functions \cite{alexa2003computing, ohtake2005multi, kazhdan2006poisson, kazhdan2013screened, xiong2014robust}.
Although mesh generation methods using the parametric function are robust against noise and missing data,
the generation of isomorphic meshes every time is not guaranteed.
Therefore, an additional remeshing process must be applied to the mesh models represented by parametric functions to obtain an isomorphic mesh.

Deformable surface meshes \cite{moore2007survey, miyauchi2018fast} can generate an isomorphic mesh by deforming a reference mesh to fit a point cloud.
However, the shape-recovery accuracy of deformable surface meshes tends to be sensitive to the initial shape of the reference mesh, as well as the noise and missing points in the target point cloud.

\subsection{Neural network-based methods}
Studies have recently been conducted on surface reconstruction from point clouds using neural networks.
These studies can be classified into two types according to the need for additional data, except the input point cloud.
Data-driven methods \cite{yu2018pu, li2019lbs, groueix2018, chen2019learning, mescheder2019occupancy, peng2020convolutional, park2019deepsdf, yang2021deep} require additional data, whereas data-free methods \cite{williams2019deep, hanocka2020point2mesh} use only the input data and no additional data.

Among data-driven methods, 
a linear-blend skinning (LBS) autoencoder \cite{li2019lbs} was proposed to generate a surface mesh by fitting a reference mesh to a target point cloud with or without missing parts.
However, the LBS autoencoder assumes that the initial shape of the reference mesh is similar to that of the target point cloud.
Therefore, for each object class, the LBS autoencoder requires an initial reference mesh with the representative or average shape of the class.

AtlasNet \cite{groueix2018} outputs a collection of local surfaces estimated from an input point cloud.
To obtain the local surfaces,  the point cloud is divided into multiple local regions.  
For each region, a multilayer perceptron (MLP) outputs a 3D point on the local surface mapped from a point on a 2D plane.
Moreover, AtlasNet generates a surface mesh by training a map from a unit spherical surface mesh to the point cloud without dividing the point cloud.
This indicates that AtlasNet generates an isomorphic mesh.
However, AtlasNet only deals with complete point clouds with no noise or missing parts.

Multiple data-driven methods represent the shape of a target object with implicit functions by a neural network, including IM-NET \cite{chen2019learning}, occupancy networks \cite{mescheder2019occupancy, niemeyer2019occupancy}, convolutional occupancy networks \cite{peng2020convolutional}, and DeepSDF \cite{park2019deepsdf, duan2020curriculum, yang2021deep, yao20213d}.
In the first three networks, the points are sampled from the 3D bounding volume covering the input point cloud. 
Using the sampled points, the network generates the implicit functions that binary classify each sampled point inside or outside the target object. 
On the contrary, DeepSDF estimates the implicit function using signed distance fields, which represent the signed distance from the surface of the target object.
However, because these networks using implicit functions output a voxel model, we need additional remeshing processes to convert the voxel model into an isomorphic mesh.

Unlike data-driven methods, data-free methods, including DGP \cite{williams2019deep} and point2mesh \cite{hanocka2020point2mesh}, enable the recovery of various shapes with no additional data except the input data.
The DGP generates a dense and
smooth point cloud from the input
point cloud with or without noise.
However, post-processing is required to generate an isomorphic mesh from the point cloud obtained by DGP.

Like the deformable model, point2mesh generates a surface mesh from the target point cloud by deforming a reference mesh for the target point cloud.
For this deformation, point2mesh learns typical shapes frequently shown in the target point cloud.
The learning process makes point2mesh robust against random noise and defects.
To improve the shape-recovery accuracy of the output mesh, point2mesh introduces a triangular patch subdivision process.
However, the subdivision process leads to a difference in the mesh structure of the generated model according to each input datum.
This indicates that the point2mesh does not guarantee the generation of isomorphic meshes represented by a common mesh structure.

%% file: method.tex
\section{iMG}
\label{sec:img}
In iMG, the target object has a genus zero closed surface. 
The input of iMG is the point cloud of the target object.
Here, the point cloud is considered to sometimes include noise and missing parts.
The reference mesh used in iMG is a spherical surface mesh generated by subdividing an icosahedron.
Using the point cloud and reference mesh, iMG generates an isomorphic mesh of the target object by deforming the reference mesh that recovers the shape of the target object.

The architecture of iMG is based on a DGP composed of MLPs.
The input point cloud when applying DGP is divided uniformly into multiple local regions once.
For each local region, 
the MLP is optimized to map
points on a 2D plane to 3D points 
on the surface of the local region.
In iMG, the MLP is optimized to map the 3D points on the reference mesh to the 3D points on the surface of the target object.
Therefore, the parameters of the optimized MLP are regarded as the parameters of the mapping function.

In general, a complex mapping function 
with many parameters has a high degree of shape representation.
Therefore, a high possibility exists that the reference mesh deformed by the complex mapping function fits the input point cloud of the target object with complex shapes.
On the contrary, 
the architecture of MLP becomes complicated according to the complexity of the mapping function.
More specifically,
to approximate the complex mapping function,
we require an MLP with many parameters.
However, 
the training of such complex MLPs sometimes fails because all the optimal parameters in the mapping function are difficult to determine.

Considering these, one approach for recovering the object shape using the mapping function is to divide the point cloud into local clouds while satisfying the condition that 
the shape of each local cloud
can be represented by 
a simpler function with few parameters.
Furthermore, iMG employs this approach by dividing the reference mesh into local meshes. 
Each local mesh is deformed to reconstruct the shape of its corresponding local point cloud.
When using the aforementioned approach, to obtain an isomorphic mesh with acceptable accuracy, we need to determine a suitable correspondence between the local meshes and the local clouds.
However, because the initial shape of the reference mesh is different from that of the point cloud, determining a suitable correspondence before deformation is difficult.
Therefore, iMG introduces a three-step strategy of deforming the reference mesh.

First, iMG performs an approximate global deformation of the reference mesh to fit the input point cloud.
Second, based on the correspondence between the deformed reference mesh in the first step and the input point cloud, the reference mesh and input point cloud are divided into a fixed number of local meshes and local clouds, respectively.
Each local mesh is deformed closer to its corresponding local cloud.
Third, the input point cloud is re-divided into local clouds according to the local shape of the input point cloud. 
Moreover, the deformed reference mesh in the second step is divided into local meshes according to the distance to the local clouds.
Each of the local meshes is further deformed to become increase its closeness to its corresponding local cloud.
The three-step strategy generates an isomorphic mesh with high shape-recovery accuracy while preventing large self-intersection of the model.

Given the initial reference mesh $\bm R^{(0)}$ and input point cloud $\bm \Pi$,
the algorithm of our iMG is described as follows (Fig. \ref{fig:method}):
\begin{enumerate}[Step 1]
\item {\bf Global mapping:} $\bm R^{(0)}$ is deformed using an MLP to approximately fit $\bm \Pi$, and the deformation result is called the first reference mesh $\bm R^{(1)}$.
\item {\bf Coarse local mapping:} $\bm R^{(1)}$ and $\bm \Pi$ are divided into 32 local meshes and their corresponding 32 local clouds. 
For each local mesh, an MLP is trained to fit the local mesh to its corresponding local cloud. 
After deforming all the local meshes, the second reference mesh $\bm R^{(2)}$ is obtained by integrating the 32 deformed local meshes.
\item {\bf Fine local mapping:} $\bm R^{(2)}$ and $\bm \Pi$ are divided into multiple local meshes and their corresponding local clouds based on the distribution of $\bm \Pi$. 
For each local mesh, an MLP is trained to fit the local mesh to its corresponding local cloud. 
The final resulting mesh $\bm R^{(3)}$ is obtained by integrating the deformed local meshes.
\end{enumerate}

\subsection{Reference mesh}
\label{sec:reference_jesh_jodel}
\begin{figure}[tb]
\begin{center}
	\includegraphics[clip, width=0.3\columnwidth]{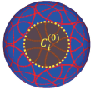}
\caption{Thirty-two local meshes illustrated on the reference mesh $\bm R^{(0)}$. The red line shows the boundary of the local meshes. The region enclosed by the yellow dotted line shows a local mesh $\bm C^{(0)}_i$}
\label{fig:division32}
\end{center}
\end{figure}

The initial reference mesh $\bm R^{(0)}$ consists of the set $\bm {V}^{(0)}$ of vertices and set $\bm T$ of triangular patches:
\begin{equation}
\bm R^{(0)} = \{\bm V^{(0)}, \bm T \} .
\end{equation}
The vertex in $\bm {V}^{(0)}$ is represented by the vector of its 3D coordinates.
The patch in $\bm T$ is a set of indices of the three vertices constituting the patch.
The initial reference mesh $\bm R^{(0)}$ is generated by subdividing an icosahedron with a circumscribed sphere of radius 1 into an arbitrary resolution.
Subsequently, $\bm R^{(0)}$ is divided into 32 local meshes (Fig. \ref{fig:division32}) to train the MLPs in coarse local mapping as follows.

We select 32 vertices, called anchor vertices, from $\bm {V}^{(0)}$.
Twelve vertices of the icosahedron in $\bm {V}^{(0)}$ are used as anchor vertices.
Moreover, using the 20 triangular elements of the icosahedron, we select the closest vertex to the line passing through the origin of the icosahedron and centroid of each triangular element.
The selected 20 vertices are also used as anchor vertices.

Second, $\bm R^{(0)}$ is divided into 32 local meshes centered at the 32 anchor vertices.
For each anchor vertex,
the vertices and patches of $\bm R^{(0)}$ are collected such that the geodesic distance from the anchor on $\bm R^{(0)}$ is less than the threshold $\tau_a$.
The set of the collected vertices and patches is defined as local mesh $\bm C^{(0)}_i$ ($i=1, 2, ... , 32$).
Since we allow the division of $\bm R^{(0)}$ such that there is an overlapping area between two or three adjacent local meshes,
the vertices of $\bm V^{(0)}$ in the overlapping area are included in multiple local meshes.
Therefore, $\bm R^{(0)} = \bm C^{(0)}_1 \bigcup ... \bigcup \bm C^{(0)}_{32}$.
The threshold $\tau_a$ used in the division of $\bm R^{(0)}$ is set such that all vertices in $\bm V^{(0)}$ are always included in at least one of the local meshes.
Through preliminary experiments, we set $\tau_a$ to $0.55$ here.
Moreover, during deformation,
the positional coordinates of
the vertices of the reference mesh
are changed while maintaining the mesh structure of the reference mesh.
Accordingly, after Steps 1 and 2 of the iMG algorithm,
the positional coordinates of
the vertices of the local mesh $\bm C^{(0)}_i$ are updated using the deformation result obtained in each step.

Coarse local mapping deforms local meshes independently.
As mentioned earlier, overlapping areas exist between the adjacent local meshes.
Therefore, the deformation of the entire reference mesh is obtained by integrating the deformed local meshes into a single smooth surface mesh.
To achieve smooth mesh integration, we define the weight of the vertex, the magnitude of which indicates the importance of the vertex for integration.
In the proposed method, 
the vertex is regarded as important for the integration
when the geodesic distance 
from the vertex in
the local mesh $\bm C^{(0)}_i$ to the
anchor vertex $\bm a_i$ of $\bm C^{(0)}_i$ is short.
Based on this concept, the weight $w^{C}_i$ of vertex $\bm v^{(0)}$ in $\bm C^{(0)}_i$ is calculated using the angle between $\bm a_i$ and $\bm v^{(0)}$, which represents the geodesic distance between them on a spherical surface:
\begin{equation}
w^{C}_i = 1 - \frac{1}{\tau_a}\arccos\left(\frac{\bm v^{(0)} \cdot \bm a_i}{\|\bm v^{(0)} \| \|\bm a_i \|}\right) ,
\end{equation}
where $\|\bm v \|$ represents the L2 norm of vector $\bm v$.
Here, weight $w^{C}_i$ is calculated once before starting the iMG algorithm and the calculated weights are
used during the algorithm.

\begin{figure}[tb]
\begin{center}
	\includegraphics[clip, width=0.8\columnwidth]{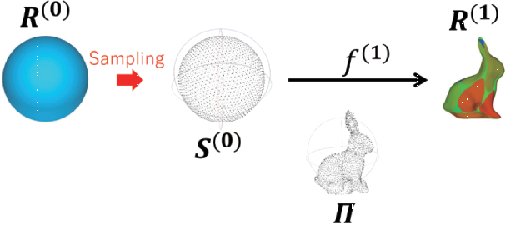}
\caption{Overview of global mapping}
\label{fig:global_mapping}
\end{center}
\end{figure}

\subsection{Global mapping}
Fig. \ref{fig:global_mapping} shows that in global mapping, the initial reference mesh $\bm R^{(0)}$ is deformed to approximately fit the input point cloud $\bm \Pi$ by training an MLP.
Here, $\bm \Pi$ is aligned in advance, such that $\bm \Pi$ is enclosed by $\bm R^{(0)}$.
The MLP in iMG uses the 3D coordinates of the points obtained from $\bm R^{(0)}$ as its input.

The MLP used in the global mapping consists of six layers: one input, four intermediate, and one output. 
The number of nodes in each layer is 3-128-256-512-512-3.
Instead of the vertices in $\bm R^{(0)}$,
we use 3D points sampled uniformly from $\bm R^{(0)}$ such that the number of sampled points is the same as that of the points in $\bm \Pi$.
Point sampling is performed using the Fibonacci sphere algorithm.
Here, the further away from $\bm \Pi$ the sampled points are, the higher the degree of freedom of movement of the points during mesh deformation.
A high degree of point movement freedom leads to flexible mesh deformation, thereby resulting in the possibility of fitting $\bm R^{(0)}$ to $\bm \Pi$.
To achieve this, 
$\bm R^{(0)}$ is enlarged by $\beta_1$ times before sampling.
Then, we sample points from the surface of the enlarged $\bm R^{(0)}$.
Scale factor $\beta_1$ is determined for each target object through preliminary experiments.
Using the set $\bm {S}^{(0)}$ of sampled points, the input of the MLP is the 3D coordinates of the sampled points in $\bm {S}^{(0)}$.
The output of the MLP is the 3D coordinates of the points after deformation.

MLP is optimized to approximate the mapping function ${f^{(1)}}$
from point $\bm s^{(0)} \in \bm {S}^{(0)}$
to its corresponding point $\bm p \in \bm \Pi$.
Using the mapping function ${f^{(1)}}$, when $\bm s^{(0)}$ is given as the input of the MLP, the output of the MLP is expressed by
${f^{(1)}}(\bm {s}^{(0)})$.
Similar to DGP, the loss function of an MLP is expressed as the mean squared error (MSE) loss between $\bm {s}^{(0)}$ and its corresponding point $\bm p$:
\begin{equation}
L^{(1)}(\bm {S}^{(0)}, \bm \Pi) = \frac{1}{\lvert\bm \Pi \rvert} \sum_{\bm {s}^{(0)} \in \bm {S}^{(0)}, \bm p \in \bm \Pi} \| f^{(1)}(\bm {s}^{(0)}) - \bm p \| ,
\label{eq:loss1}
\end{equation}
where $\lvert \bm \Pi \rvert$ represents the number of the points in the set $\bm \Pi$.
For each sampled point $\bm {s}^{(0)}$, the corresponding point $\bm p$ is determined using the Sinkhorn regularised distance \cite{cuturi2013sinkhorn} between $\bm {S}^{(0)}$ and $\bm \Pi$.
The correspondence is redefined whenever the parameters of the MLP are updated.

After the global mapping,
we obtain the first reference mesh
$\bm R^{(1)}$ by changing $\bm v^{(0)} \in \bm V^{(0)}$
into the vertex $\bm v^{(1)} = {f^{(1)}}(\beta_1 \bm v^{(0)})$. 
A set of $\bm v^{(1)}$ is denoted by $\bm V^{(1)}$.

Here, depending on the initial values of the MLP parameters, the normal vectors at the vertices in $\bm V^{(1)}$ may be flipped or twisted during MLP training.
To prevent this, a normal penalty is calculated for several epochs of MLP training.
The normal penalty is defined as the average inner product between the normal vectors at each point before and after mapping.
If the penalty is less than or equal to zero, we consider that the deformed reference mesh contains flipped or twisted parts.
The parameters of the MLP are randomly reinitialized and retrained.

\subsection{Coarse local mapping}
\label{sec:coarse_local_mapping}
\begin{figure}[tb]
\begin{center}
	\includegraphics[clip, width=1.0\columnwidth]{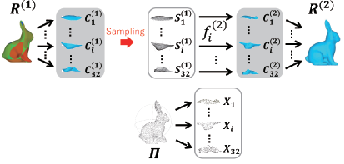}
\caption{Overview of coarse local mapping}
\label{fig:local_mapping}
\end{center}
\end{figure}

Coarse local mapping consists of three steps (Fig. \ref{fig:local_mapping}). 
The first step divides the input point cloud $\bm \Pi$ into 32 local clouds $\bm X_i$ corresponding to the 32 local meshes $\bm C^{(1)}_i$.
Here, $\bm C^{(1)}_i$ is obtained by replacing the vertices in $\bm C^{(0)}_i$ with their corresponding vertices in $\bm R^{(1)}$.
In the division of $\bm \Pi$, for each point $\bm p \in \bm \Pi$,
we find the closest vertex 
to $\bm p$ from the first reference mesh $\bm R^{(1)}$
and regard it as the corresponding vertex of $\bm p$.
Using the correspondence between the points in $\bm \Pi$ and the vertices in $\bm R^{(1)}$,
the local cloud $\bm X_i$ 
is determined by 
collecting the points whose corresponding vertices are included in $\bm C^{(1)}_i$.

In the second step, each local mesh $\bm C^{(1)}_i$ is deformed to fit its corresponding local cloud $\bm X_i$.
Finally, the second reference mesh $\bm R^{(2)}$ is obtained by integrating 32 deformed local meshes $\bm C^{(2)}_i$.
The following describes the details of the second and third steps of coarse local mapping.

\subsubsection{Deformation of each local mesh}
\label{sec:doelmr}
In the local mesh deformation, the MLP is optimized to approximate the mapping function from the local mesh $\bm C^{(1)}_i$ to the corresponding local cloud $\bm X_i$ (Fig. \ref{fig:local_mapping}).
Before deformation, the coordinate system of $\bm C^{(1)}_i$ is transformed from the global coordinate system of the reference mesh to a local coordinate system, the origin of which coincides with the centroid of $\bm C^{(1)}_i$.
Moreover, $\bm X_i$ is transformed into the local coordinate system by applying the transformation of $\bm C^{(1)}_i$ from the global coordinate system to the local coordinate system.

Next, using the same method of generating sampled points in the global mapping (Step 1 of the iMG algorithm), $\bm C^{(1)}_i$ is enlarged by a scale factor $\beta_2$.
3D points are sampled uniformly from the
surface of the enlarged $\bm C^{(1)}_i$ such that the number of sampled points is the same as that of $\bm X_i$. 
The sampling is performed using Lloyd's algorithm.
The scale factor, $\beta_2$, is determined for each target object through preliminary experiments.
The set $\bm {S}^{(1)}$ of the sampled points is used as the training dataset for the MLP during coarse local mapping.

The MLP is optimized to approximate
the mapping function $f^{(2)}_i$ from the sampled point $\bm s^{(1)} \in \bm S^{(1)}_i$ to its corresponding point in $\bm X_i$.
Therefore, the output of the MLP is represented by
${f}^{(2)}_i(\bm s^{(1)})$.
The architecture of the MLP is the same as that of the MLP used in global mapping.
The loss function is represented by Eq. (\ref{eq:loss1}).
Before training the
MLP, Sinkhorn regularised distance is employed to determine the correspondence between $\bm X_i$ and $\bm {S}^{(1)}_i$. 
The correspondence is used with no updates during training to
prevent self-intersections between adjacent local meshes.

As mentioned in Section \ref{sec:reference_jesh_jodel},
an overlapping area exists between adjacent local meshes,
and their corresponding adjacent local clouds overlap.
Therefore, multiple sampled points correspond to a point in the overlapping area of the local clouds.
The outputs of the different MLPs corresponding to the point
often have different positions.
This difference leads to the discontinuity between the adjacent local surfaces estimated using the MLPs.
To fuse the adjacent surfaces smoothly into one surface, we must train the MLPs while minimising the difference between multiple outputs.
To achieve this, MLP training consists of two steps based on the concept of DGP.

First, each MLP is trained using the local cloud $\bm X_i$.
After finishing the training, 
we calculate the outputs of the trained MLPs using the sampled points in set $\bm S^{(1)}_i$. 
The outputs of the MLPs are used as new positional coordinates of
the points in $\bm \Pi$ to retrain the MLPs.

Here, in the case of the point in the overlapping area, 
there are several positional coordinates of the point obtained using different MLPs.
The multiple output points are integrated into a new point by calculating the weighted average of the positional coordinates.
The weight of the output point is determined as follows.
Since a point in $\bm X_i$ correspond to a vertex in the local mesh $\bm C^{(1)}_i$, 
the output point of the MLP corresponding to the point in $\bm X_i$ is regarded as the corresponding point of the vertex. 
Using this correspondence,
we use as the weight of the output point
the weight $w^{C}_i$ of its corresponding vertex in $\bm C^{(1)}_i$.

Using the new positional coordinates of the points in $\bm \Pi$, the MLPs are re-trained using the loss function (Eq. (\ref{eq:loss1})).
After additional training,
we obtain the deformed local mesh $\bm C^{(2)}_i$ by replacing the 3D coordinates of the vertex $\bm v^{(1)}$ in $\bm C^{(1)}_i$
with their mapped 3D coordinates ${f^{(2)}}(\beta_2 \bm v^{(1)})$.

\subsubsection{Integration of the deformed local mesh}
\label{sec:Iodr}
The deformation result of $\bm R^{(1)}$,
called the second reference mesh $\bm R^{(2)}$,
is generated by integrating the deformed local meshes $\bm C^{(2)}_i$.
During the retraining of the MLPs, multiple MLPs are trained to map vertex $\bm v^{(1)}$ in the overlapping area to the same point.
However, the mapped points,
which are the output points of MLPs
using $\bm v^{(1)}$,
do not always have the same coordinates.
Therefore, we fuse the output points into one vertex $\bm v^{(2)}$ by calculating the weighted average of the outputs.
During the calculation, the weight $w^{C}_i$ of $\bm v^{(1)}$ is used as the weight of the output point obtained from the MLP for $\bm C^{(1)}_i$.
After the fusion, we generate the second reference mesh $\bm R^{(2)}$ by changing $\bm v^{(1)} \in \bm V^{(1)}$ into vertex $\bm v^{(2)}$.
A set of $\bm v^{(2)}$ is denoted as $\bm V^{(2)}$.

\subsection{Fine local mapping}
\begin{figure}[tb]
\begin{center}
	\includegraphics[clip, width=1.0\columnwidth]{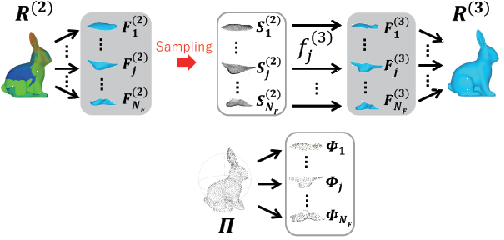}
\caption{Overview of fine local mapping}
\label{fig:fine_mapping}
\end{center}
\end{figure}

Fine local mapping consists of three steps (Fig. \ref{fig:fine_mapping}).
Based on the shape of the input point cloud $\bm \Pi$,
the first step is to divide the second reference mesh $\bm R^{(2)}$ and $\bm \Pi$ into $N_F$ local meshes $\bm F^{(2)}_j$ and $N_F$ local clouds $\bm \Phi_j$ ($j=1, ... , N_f$), respectively.
In the second step, using an MLP, each local mesh $\bm F^{(2)}_j$ is deformed to fit its corresponding local cloud $\bm \Phi_j$. 
Finally, the final reference mesh $\bm R^{(3)}$ is obtained by integrating the $N_F$-deformed local meshes.
The details of each step in the fine local mapping are described as follows.

\subsubsection{Division of the second reference mesh and input point cloud}
\label{sec:surface_division}
\begin{figure}[tb]
\begin{center}
	\includegraphics[clip, width=1.0\columnwidth]{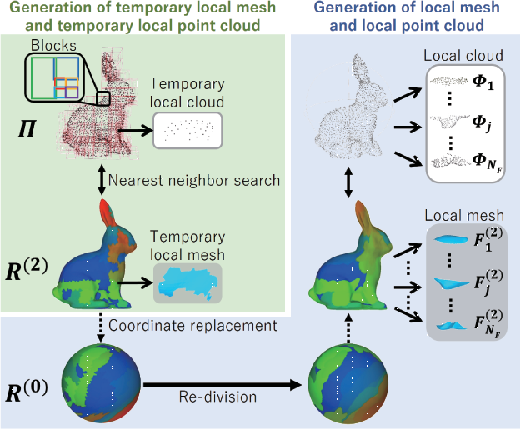}
\caption{Division of the second reference mesh and input point cloud. Different colours are assigned to each block or local mesh}
\label{fig:division_with_surface_approximation}
\end{center}
\end{figure}

The division of the second reference mesh $\bm R^{(2)}$ and input point cloud $\bm \Pi$ primarily consists of two processes. 
In the first process, the input point cloud $\bm \Pi$ is divided using a set of multiple blocks which cover $\bm \Pi$ (top-left figure in Fig. \ref{fig:division_with_surface_approximation}).
The point cloud within each block is referred to as a temporary local cloud.
Furthermore, $\bm \Pi$ is divided such that the shape of the temporary local cloud is represented by a simple function of the polynomial surface.
The use of the temporary local clouds enables the estimation of the mapping function of the temporary local cloud using the MLP with fewer parameters, as described in Section \ref{sec:img}.
Next, for each temporary local cloud,
we collect the vertex in $\bm R^{(2)}$ with the closest Euclidean distance to each point in the temporary local cloud (middle-left figure in Fig. \ref{fig:division_with_surface_approximation}).
The set of collected vertices and patches consisting of the collected vertices is called a temporary local mesh.

Here, the local mesh, which is the input of the MLP, is assumed to be a continuous area with no holes.
However, depending on the shape of the second reference mesh, $\bm R^{(2)}$, the temporary local mesh is often discontinuous and/or has holes.
Therefore, in the second process, using the initial reference mesh $\bm R^{(0)}$ and temporary local meshes (bottom-left and right figures in Fig. \ref{fig:division_with_surface_approximation}),
$\bm R^{(2)}$ is re-divided into $N_F$ continuous areas, called local meshes (middle-right figure in Fig. \ref{fig:division_with_surface_approximation}).
Moreover, for each of the obtained local mesh in $\bm R^{(2)}$, its corresponding local cloud is re-defined as the set of closest points in $\bm \Pi$ to the vertices in the local mesh
(top-right figure in Fig. \ref{fig:division_with_surface_approximation}).
The details of each process are described as follows.

\paragraph{Generation of a temporary local mesh}
To divide the input point cloud $\bm \Pi$ into temporary local points, we use a bounding box containing $\bm \Pi$,
which is a set of $n_{1} \times n_{1} \times n_{1}$ ($n_{1} = 10$ in our method) blocks of the same size.
For each block, 
the fitting error ${E}$ of the points in the block is defined as
the maximum distance between each point $\bm p$ and a point on the polynomial surface, where the $x$ and $y$ coordinates are the same as those of $\bm p$.
Here, in our method, a fifth-order polynomial is used as a function of the polynomial surface. 
When ${E}$ is larger than a threshold $\tau_e$, the block is divided evenly into $n_{2}\times n_{2}\times n_{2}$ ($n_{2} = 2$ in our method) sub-blocks.
If ${E}$ for even one of the obtained sub-blocks continues to be greater than $\tau_e$, $n_2$ is replaced by $n_2 + 1$, and the block is divided into sub-blocks using the new $n_2$. 
These processes are repeated until ${E}$ becomes less than $\tau_e$.
After these processes, all subblocks are regarded as blocks.

After applying the division process to all the blocks, many blocks are generated.
The following processes in the fine local mapping use the generated blocks. 
Therefore, the computational time for fine local mapping increases according to the number of blocks.
To reduce the computational time while maintaining shape-recovery accuracy, adjacent blocks are merged into a new block.

For two arbitrary adjacent blocks, 
we find the polynomial surface that fits the shape of the points in the two blocks.
If the fitting error ${E}$ for the points in the two blocks is less than the threshold $\tau_e$, the two sub-blocks are merged into one block.
The process of merging adjacent blocks is repeated until there are no pairs of adjacent blocks 
with ${E}$ less than $\tau_e$. 

After merging, all adjacent blocks are enlarged
by multiplying the length of each side of the blocks by $1.01$ in our method (inside the black frame in the top-left figure in Fig. \ref{fig:division_with_surface_approximation}). 
This is because one arbitrary patch in $\bm R^{(2)}$ is always included in at least one temporary local mesh.
In contrast, owing to block enlargement, there is
always an overlapping area among arbitrary adjacent temporary
local meshes.
Using these blocks, the input point cloud $\bm \Pi$ is divided into $N_F$-temporary point clouds.
For each block, its corresponding temporary point cloud is obtained by collecting points within the block.
Moreover, for each temporary point cloud, the corresponding temporary local mesh is obtained by collecting the closest vertices in $\bm R^{(2)}$ to the points in the temporary point cloud.

\paragraph{Generation of a local mesh and local point cloud}
First, the positional coordinates of the vertices in the temporary local meshes are replaced with those of their corresponding vertices in the initial reference mesh $\bm R^{(0)}$.
This replacement is regarded as the division of the initial reference mesh
(the spherical surface) by using the temporary local meshes.
Next, we calculate the smallest elliptic cylinder that includes all vertices in the temporary local mesh.
Part of the initial reference mesh $\bm R^{(0)}$ within the elliptic cylinder is defined as a local mesh.
By applying the elliptic cylinder calculation to all temporary local meshes, $\bm R^{(0)}$ is divided into $N_F$-local meshes $\bm F^{(0)}_j$.

Finally, we obtain the local mesh $\bm F^{(2)}_j$ of the second reference mesh $\bm R^{(2)}$
by replacing the coordinates
of the vertices in $\bm F^{(0)}_j$
with those of their corresponding
vertices in $\bm R^{(2)}$ (Fig. \ref{fig:division_with_surface_approximation}).
Moreover, 
for each point $\bm p$ in the input cloud $\bm \Pi$,
we find the closest vertex 
to $\bm p$ from $\bm R^{(2)}$
and regard this as the corresponding vertex of $\bm p$.
Using the correspondence between the points in $\bm \Pi$ and the vertices in $\bm R^{(2)}$,
the local cloud $\bm \Phi_j$ 
is determined by 
collecting the points whose corresponding vertices are included in $\bm F^{(2)}_j$.

Here, because all the blocks overlap, an overlapping area exists between the adjacent elliptic cylinders. 
Therefore, the vertices in the overlapping area are included in the multiple local meshes.
Similar to the coarse local mapping;
in the fine local mapping,
the MLP outputs the deformation of each local mesh.
Then, 
as described in Section \ref{sec:doelmr}, in the case of the point in the overlapping area, several positional coordinates exist of the point obtained
from the different MLPs.
Accordingly, to achieve a smooth mesh integration, we define a weight $w^{F}_j$ of the vertex in $\bm F^{(2)}_j$, the magnitude of which indicates the importance of the vertex for integration.
For each vertex in $\bm F^{(2)}_j$, we define $w^{F}_j$ as the geodesic distance on $\bm R^{(0)}$ from the centroid of $\bm F^{(0)}_j$ to the vertex in  $\bm F^{(0)}_j$.
To ensure that the weights around the boundary of $\bm F^{(0)}_j$ are 0, the distance is normalised using the lengths of the long and short sides of an approximated ellipse for $\bm F^{(0)}_j$.

\subsubsection{Deformation of the local mesh}
\label{sec:doelm2}
In the fine local mapping, the MLP is optimized to approximate the mapping function from the local mesh $\bm F^{(2)}_j$ to the corresponding local cloud $\bm \Phi_j$.
Before the deformation, the coordinate system of $\bm F^{(2)}_j$ is transformed from the global coordinate system of the reference mesh to a local coordinate system, the origin of which coincides with the centroid of $\bm F^{(2)}_j$.
Using the coordinate transformation of $\bm F^{(2)}_j$, $\bm \Phi_j$ is converted into the local coordinate system.

The training dataset for MLP consists of two sets of 3D coordinates.
The first set is the set of sampled points using $\bm F^{(2)}_j$.
The sampling method is the same as that for generating the sampled points in the coarse local mapping (Step 2 of the iMG Algorithm).
After $\bm F^{(2)}_j$ is enlarged by scaling factor $\beta_3$, the 3D points are sampled uniformly from
the surface of the enlarged $\bm F^{(2)}_j$ such that the number of sampled points is the same as that of $\bm \Phi_j$. 
This sampling is performed using Lloyd's algorithm.
The scale factor, $\beta_3$, is determined for each target object through preliminary experiments.
Set $\bm S^{(2)}_j$ of the sampled points is used as the first set.
Subsequently, the local cloud $\bm \Phi_j$ is used as the ground truth for $\bm S^{(2)}_j$.

Here, when the number $\lvert \bm \Phi_j \rvert$ of the points in $\bm \Phi_j$ is large, the MLP can be optimized efficiently using only $\bm S^{(2)}_j$.
However, because the rugged area cannot be represented by one polynomial surface, the area is divided into multiple local clouds with a small number of points.
This indicates that many local clouds with a small number of points are generated around the rugged area. 
Therefore, the use of only $\bm S^{(2)}_j$ is sometimes insufficient for training the MLP.

To stably train the MLP using even a small number of points in $\bm \Phi_j$, 
the set $\bm V^F_j$ of vertices in $\bm F^{(2)}_j$ is added to the training dataset as the second set.
Moreover, the set $\bm S^p_j$ is generated using the polynomial surface
for $\bm \Phi_j$ and used as the ground truth of $\bm V^F_j$. 
Here, the polynomial surface is determined in the same way as for the block subdivision discussed in
Section \ref{sec:surface_division}.
The 3D points are randomly sampled from the polynomial surface such that the number of sampled points is the same as that of the vertices in $\bm V^F_j$.
The set of the sampled points is used as the set $\bm S^p_j$.

The MLP is optimized to approximate
the mapping function ${f}^{(3)}_j$ from the sampled point $\bm s^{(2)} \in \bm S^{(2)}_j$ to its corresponding point $\bm p \in \bm \Phi_j$ and from $\bm v^{(2)} \in \bm V^F_j$ to $\bm s^p \in \bm S^p_j$.
Therefore, the output of the MLP using $\bm s^{(2)}$ or $\bm v^{(2)}$ is represented by
${f}^{(3)}_j(\bm s^{(2)})$ or ${f}^{(3)}_j(\beta_3 \bm v^{(2)})$.
The architecture of the MLP is the same as that of the MLP used in the global mapping.
The loss function is expressed as
\begin{equation}
\begin{split}
	{L}^{(3)}(\bm S^{(2)}_j, &\bm \Phi_j, \bm V^F_j, \bm S^p_j) \\
& = \frac{1}{\lvert \bm \Phi_j \rvert} \sum_{\bm s^{(2)} \in \bm S^{(2)}_j \bm p \in \bm \Phi_j} \| {f^{(3)}_j}(\bm s^{(2)}) - \bm p \| \\
	& + \frac{\alpha }{\lvert \bm V^F_j \rvert} \sum_{\bm v^{(2)} \in \bm V^F_j, \bm s^p \in \bm S^p_j} \| {f^{(3)}_j}(\beta_3 \bm v^{(2)}) - \bm{s}^p \| .
\label{eq:loss3}
\end{split}
\end{equation}
The first term of Eq. (\ref{eq:loss3}) represents an input point cloud-based loss, which evaluates the mapping accuracy from the sampled points on the reference mesh $\bm R^{(2)}$ to the input point cloud.
The second term of Eq. (\ref{eq:loss3}) represents a reference mesh-based loss which evaluates the mapping accuracy from the reference mesh to sampled points $\bm{s}^p$ on a polynomial surface.
Parameter $\alpha$ is a weight coefficient that determines whether the input point cloud-based loss or the reference mesh-based loss is emphasised.
Based on preliminary experiments, we set $\alpha=0.5$.
Before training the
MLP, the Sinkhorn regularised distance is employed to determine the correspondence between $\bm \Phi_j$ and $\bm S^{(2)}_j$ and between $\bm S^p_j$ and $\bm V^F_j$. 
Similar to the coarse local mapping, the correspondence is used with no updates during training.

The MLP is trained in the same way as training the MLP in the coarse local mapping (Section \ref{sec:doelmr}).
First, for each local cloud $\bm \Phi_j$,
the MLP is trained using $\bm \Phi_j$.
After finishing the training, 
we calculate the output of the trained MLPs when the sampled points in $\bm S^{(2)}_j$
are given as the inputs of the MLPs. 
The output of the MLPs is used as new coordinates of
the points in $\bm \Pi$ to retrain the MLPs.
In the case of a point in the overlapping area, 
its multiple
output points are integrated into one new point by calculating the weighted average of the positional coordinates.
Here, as the weight of the output point, we use weight $w^{F}_j$ of its corresponding vertex in $\bm F^{(2)}_j$.
The MLPs are re-trained using the new positional coordinates of the points in $\bm \Pi$.
For the additional training, we set $\alpha = 0$ in Eq. (\ref{eq:loss3})
to increase the emphasis on mapping from the sampled points on the reference mesh to the input point cloud.
After additional training,
we obtain deformed local mesh $\bm F^{(3)}_j$ by replacing the 3D coordinates of the vertex $\bm v^{(2)}$ in $\bm F^{(2)}_j$
with their mapped 3D coordinates ${f^{(3)}}(\beta_3 \bm v^{(2)})$.

\subsubsection{Integration of the deformed local mesh}
\label{subsubsec:integration}
Similar to the method used in coarse local mapping,
the entire deformation result of $\bm R^{(2)}$,
called the final reference mesh $\bm R^{(3)}$,
is obtained by integrating the deformed local meshes $\bm F^{(3)}_j$.
However, the mapped points,
which are the output points of MLPs
using vertex $\bm v^{(2)}$ in the overlapping area,
do not always have the same coordinates.
Considering this, we fuse the output points into one vertex $\bm v^{(3)}$ by calculating the weighted average of the outputs.
The weight of the output point obtained using the MLP for $\bm F^{(2)}_j$ is the weight $w^{F}_j$ of $\bm v^{(2)}$.

In fine local mapping, more than three local meshes share the same overlapping areas. 
This means that more than three output points are obtained using the MLPs for the local meshes.
In this case, we select only the output points
with high weights to reduce the influence of the output point with low shape-recovery accuracy.
The weighted average is calculated using the selected output points.
Here, we select the output points with the first third-highest weight.
After fusion, we obtain the final reference mesh $\bm R^{(3)}$ by changing $\bm v^{(2)} \in \bm V^{(2)}$ into vertex $\bm v^{(3)}$.

%% file: experiment.tex
\begin{figure}[t]
\begin{center}
	\subfloat[][]{\includegraphics[clip, width=0.25\columnwidth]{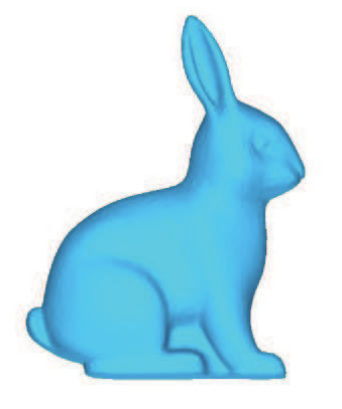}} \quad
	\subfloat[][]{\includegraphics[clip, width=0.2\columnwidth]{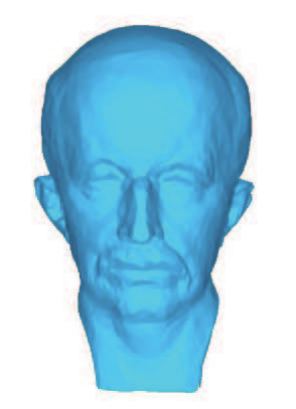}} \quad
	\subfloat[][]{\includegraphics[clip, width=0.25\columnwidth]{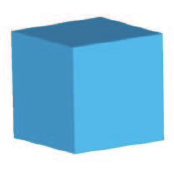}}
\caption{Ground truths: (a) rabbit, (b) Max Planck, (c) box}
\label{fig:simulation_mesh}
\end{center}
\end{figure}

\begin{figure}[t]
\begin{center}
	\subfloat[][]{\includegraphics[clip, width=0.25\columnwidth]{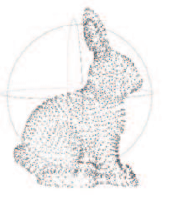}} \quad
	\subfloat[][]{\includegraphics[clip, width=0.3\columnwidth]{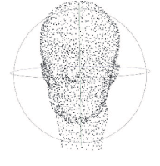}} \quad
	\subfloat[][]{\includegraphics[clip, width=0.25\columnwidth]{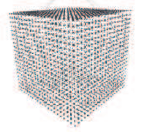}}
\caption{Simulation data without noise: (a) the rabbit, (b) Max Planck, and (c) the box models}
\label{fig:simulation_points}
\end{center}
\end{figure}

\begin{figure}[t]
\begin{center}
	\subfloat[][]{\includegraphics[clip, width=0.25\columnwidth]{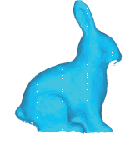}} \quad
	\subfloat[][]{\includegraphics[clip, width=0.2\columnwidth]{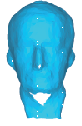}} \quad
	\subfloat[][]{\includegraphics[clip, width=0.25\columnwidth]{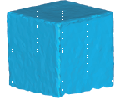}}
\caption{Measurement meshes obtained by scanning real models: (a) the rabbit, (b) Max Planck, and (c) the box models}
\label{fig:mesh_model_of_real_model}
\end{center}
\end{figure}

\begin{figure}[t]
\begin{center}
	\subfloat[][]{\includegraphics[clip, width=0.25\columnwidth]{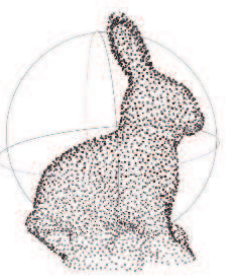}} \quad
	\subfloat[][]{\includegraphics[clip, width=0.28\columnwidth]{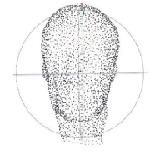}} \quad
	\subfloat[][]{\includegraphics[clip, width=0.25\columnwidth]{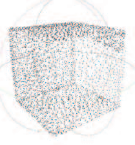}}
\caption{Measurement data: (a) the rabbit, (b) Max Planck, and (c) the box models}
\label{fig:scanned_points}
\end{center}
\end{figure}

\section{Experiment}
\label{sec:experiment}
To confirm the effectiveness of the proposed iMG, we conducted experiments to generate isomorphic meshes using simulation and measurement data.
In the simulations, we used two types of simulation data: point cloud data with and without noise.
The measurement data were acquired using the TrueDepth camera of an iPhone 11 Pro (Apple Inc., USA).

AtlasNet is a well-known method that generates an isomorphic mesh from a spherical surface mesh, similar to our iMG. 
In the experiments, we compared AtlasNet with iMG, which
optimized the network each time to generate an isomorphic mesh. 
This indicates that with the same input point cloud, the shape of the generated model is different for each optimizing time.
Therefore, iMG generated its isomorphic mesh five times from each input point cloud for evaluation.
The shape-recovery accuracy of iMG from the input point cloud was evaluated using the average and standard deviation of the shape-recovery accuracy of the five generated models.
However, 
AtlasNet uses a trained network to generate an isomorphic mesh. 
The use of the trained network leads to the unique generation of an isomorphic mesh from the input point cloud. 
Therefore, AtlasNet generated an isomorphic mesh only once.
In all the experiments, a GPU (MSI Radeon RX Vega 56 Air Boost 8G OC) was used.

\begin{figure}[t]
\begin{center}
	\subfloat[][]{\includegraphics[clip, width=0.25\columnwidth]{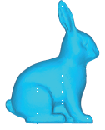}} \quad
	\subfloat[][]{\includegraphics[clip, width=0.2\columnwidth]{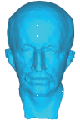}} \quad
	\subfloat[][]{\includegraphics[clip, width=0.25\columnwidth]{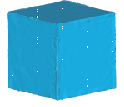}} 
\caption{Experiment1: iMG generated the isomorphic meshes of (a) the rabbit, (b) Max Planck, and (c) the box models}
\label{fig:structured_mesh_models_for_simulation}
\end{center}
\end{figure}

\begin{figure}[t]
\begin{center}
	\subfloat[][]{\includegraphics[clip, width=0.25\columnwidth]{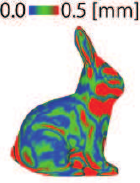}} \quad
	\subfloat[][]{\includegraphics[clip, width=0.25\columnwidth]{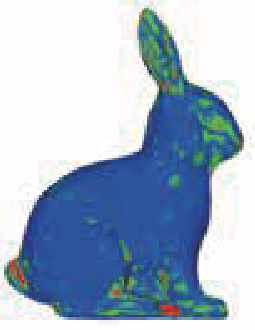}} \quad
	\subfloat[][]{\includegraphics[clip, width=0.25\columnwidth]{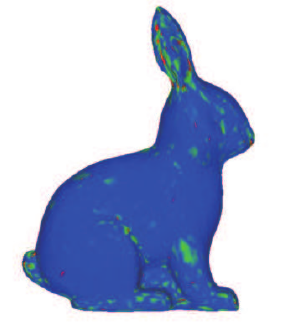}} \quad
	\subfloat[][]{\includegraphics[clip, width=0.25\columnwidth]{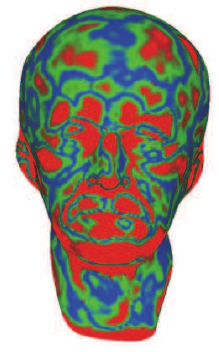}} \quad
	\subfloat[][]{\includegraphics[clip, width=0.25\columnwidth]{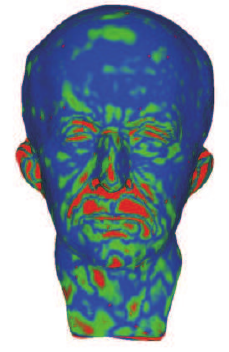}} \quad
	\subfloat[][]{\includegraphics[clip, width=0.25\columnwidth]{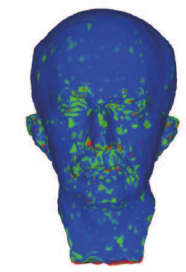}} \quad
	\subfloat[][]{\includegraphics[clip, width=0.25\columnwidth]{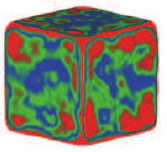}} \quad
	\subfloat[][]{\includegraphics[clip, width=0.25\columnwidth]{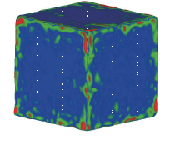}} \quad
	\subfloat[][]{\includegraphics[clip, width=0.25\columnwidth]{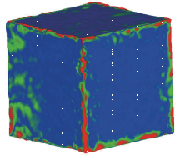}} 
\caption{Experiment1: Isomorphic meshes after (a, d, g) global mapping, (b, e, h) coarse local mapping, and (c, f, i) fine local mapping of (a--c) the rabbit, (d--f) Max Planck, and (g--i) the box models}
\label{fig:mapping_results_for_simulation}
\end{center}
\end{figure}

\begin{table}[t]
\centering
\caption{Results of Experiment 1: The second to fourth columns lists the PM distances obtained from global mapping, coarse local mapping, and fine local mapping, respectively}
\begin{tabular}{l|lll}
\hline
               & \multicolumn{3}{c}{PM distance [mm]} \\ \cline{2-4} 
           & Global      & Coarse      & Fine                \\ \hline
Rabbit     & $0.39 \pm 0.01$ & $0.12\pm 0.00$ & $0.07 \pm 0.00$      \\
Max Planck & $0.48 \pm 0.03$ & $0.20 \pm 0.02$ & $0.08 \pm 0.01$       \\
Box        & $0.39 \pm 0.02$ & $0.10 \pm 0.00$ & $0.11 \pm 0.01$      \\ \hline
\end{tabular}
\label{table:result_for_exp1}
\end{table}

\subsection{Data}
In the simulation, we used three surface meshes: a rabbit,
a replica of the face of Max Planck, and a box, as shown in Fig. \ref{fig:simulation_mesh}.
The vertices of the meshes were used as the simulation data without noise (Fig. \ref{fig:simulation_points}), whereas the original surface meshes were employed as ground truths of the isomorphic meshes generated from the simulation data.
The sizes of the three ground truths were 75.2 $\times$ 35.7 $\times$ 98.0 [mm], 72.5 $\times$ 59.2 $\times$ 98.9 [mm], and 100.0 $\times$ 100.0 $\times$ 100.0 [mm], respectively.

To obtain the measurement data, we created the three real models of the three meshes using a 3D printer. 
The shapes of the real models were measured using a TrueDepth camera.
During the measurement, the mobile phone was fixed on a tripod, whereas the real model was placed on a rotating table. 
By rotating the table, we scanned the entire surface of the model.
Through the measurement,  
we obtained the surface meshes of the real models using STL Maker (Scandy LLC, New Orleans, LA, USA).

To perform iMG with only one GPU, the number of the vertices of  
each surface mesh was reduced to approximately 2,500 by a down-sampling process. 
The surface mesh obtained using the down-sampling process was used as the measurement mesh (Fig. \ref{fig:mesh_model_of_real_model}). 
Moreover, the vertices of the measurement mesh were used as measurement data (Fig. \ref{fig:scanned_points}).
The measurement data always contained noise.
As shown in Fig. \ref{fig:mesh_model_of_real_model}, there were missing parts in the measurement data owing to the occlusion of the object.
For example, there were no points on the jaw of the rabbit, nose and neck of the Max Planck model, and bottom of the
mesh that was in contact with the rotating table during the measurement.
The ground truths (Fig. \ref{fig:simulation_mesh}) for the simulation data were employed as ground truths for the measurement data.

\subsection{Accuracy Metric}
To evaluate the shape-recovery accuracy of the isomorphic mesh, we used the point-mesh distance (PM distance)
which is the average of the two distances from the isomorphic mesh $\bm R$ to its ground truth $\bm G$ and vice versa.
Here, the distance $d(\bm v, \bm G)$ from a vertex $\bm v$ in $\bm R$ to $\bm G$ is defined by
\begin{equation}
d(\bm v, \bm G) = \min_{\gamma \in \bm \Gamma} H(\bm v, \gamma),
\label{eq:d_v}
\end{equation}
where $\bm \Gamma$ is the set of triangular patches in the ground truth $\bm G$.
Function $H(\bm v, \gamma)$ returns the length of the perpendicular line from vertex $\bm v$ to patch $\gamma$.

By contrast, the distance $d(\bm p, \bm R)$ from point $\bm p$ in $\bm G$ to $\bm R$ is defined as
\begin{equation}
d(\bm p, \bm R) = \min_{t \in \bm T} H(\bm p, t),
\label{eq:d_o}
\end{equation}
where $\bm T$ is the set of triangular patches in isomorphic mesh $\bm R$.
The PM distance $D(\bm R, \bm G)$ is defined as
\begin{equation}
D(\bm R, \bm G) = \frac{1}{2} \Bigl(\frac{\sum_{\bm v \in \bm V}d(\bm v, \bm G)}{\lvert \bm V \rvert} + \frac{\sum_{\bm p \in \bm \Pi}d(\bm p, \bm R)}{\lvert \bm \Pi \rvert} \Bigr),
\label{eq:PM}
\end{equation}
where $\bm V$ and $\bm \Pi$ are the sets of vertices and points in isomorphic mesh $\bm R$ and its ground truth $\bm G$, respectively.
The shape-recovery accuracy increases with the decrease in PM distance.

\subsection{Experiment 1: Isomorphic mesh generation using simulation data without noise}
In the simulation using the three simulation data without noise (Fig. \ref{fig:simulation_points}),
the number of vertices of the reference mesh was 36,002.
Through preliminary experiments, the hyperparameters $\beta_1$, $\beta_2$, $\beta_3$, and $\tau_e$ of iMG for each datum were determined.

The iMG generated isomorphic meshes (Fig. \ref{fig:structured_mesh_models_for_simulation}) using the three simulation data as the input point clouds.
Fig. \ref{fig:mapping_results_for_simulation} shows the isomorphic meshes after global mapping, coarse local mapping, and fine local mapping.
Table \ref{table:result_for_exp1} shows the PM distances for the three mappings.
The colour map in Fig. \ref{fig:mapping_results_for_simulation} illustrates the distance $d(\bm v, \bm G)$ from vertex $\bm v$ of $\bm R$ to $\bm G$ in Eq. (\ref{eq:d_v}).
The colour was close to blue when the value of $d(\bm v, \bm G)$ was close to zero, whereas the colour was close to red when the value was close to 0.5 [mm].

From Fig. \ref{fig:mapping_results_for_simulation} and Table \ref{table:result_for_exp1}, for the rabbit and Max Planck models, we observed that the blue region increased, whereas the PM distance decreased as the deformation progressed from global mapping to fine local mapping.
This indicated that the distance between the isomorphic mesh and its ground truth decreased.
In particular, the shape-recovery accuracy improved after coarse local mapping. 
On the contrary, the accuracy remained low in shapes with large curvature changes such as the tail, eyes, nose, and hollows at the feet of the rabbit, as well as the ears, eyes, and mouth of the Max Planck model.
However, fine local mapping improved the shape-recovery accuracy in these regions.

The PM distance after coarse local mapping of the box was the shortest among the three steps.
After fine local mapping, the red region around the edges increased slightly.

\begin{figure}[tb]
\begin{center}
	\subfloat[][]{\includegraphics[clip, width=0.25\columnwidth]{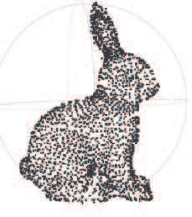}} \quad
	\subfloat[][]{\includegraphics[clip, width=0.25\columnwidth]{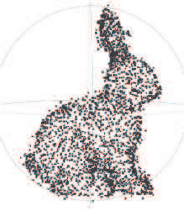}} \quad
	\subfloat[][]{\includegraphics[clip, width=0.25\columnwidth]{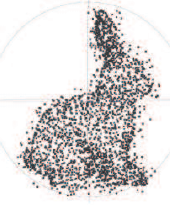}} \quad
	\subfloat[][]{\includegraphics[clip, width=0.25\columnwidth]{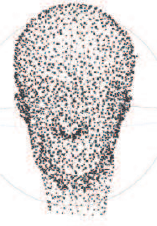}} \quad
	\subfloat[][]{\includegraphics[clip, width=0.25\columnwidth]{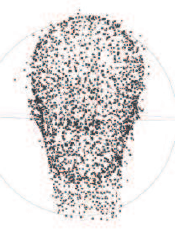}} \quad
	\subfloat[][]{\includegraphics[clip, width=0.25\columnwidth]{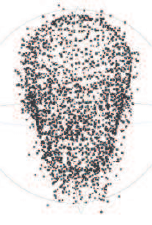}} \quad
	\subfloat[][]{\includegraphics[clip, width=0.25\columnwidth]{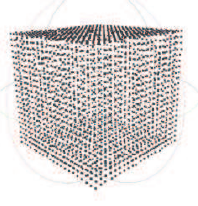}} \quad
	\subfloat[][]{\includegraphics[clip, width=0.25\columnwidth]{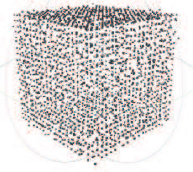}} \quad
	\subfloat[][]{\includegraphics[clip, width=0.25\columnwidth]{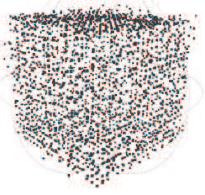}} 
	
\caption{Experiment 2: (a--c), (d--f), and (g--i) are the simulation data with noise of the rabbit, Max Planck, and the box models, respectively. From left to right, the noise parameters ($\delta$, $\sigma$) are (20, 1.0), (40, 3.0), and (60, 5.0)}
\label{fig:noised_input_points_for_simulation}
\end{center}
\end{figure}

\begin{figure}[tb]
\begin{center}
	\subfloat[][]{\includegraphics[clip, width=0.28\columnwidth]{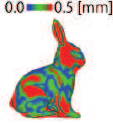}} \quad
	\subfloat[][]{\includegraphics[clip, width=0.25\columnwidth]{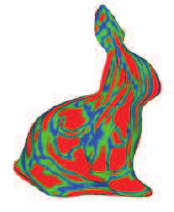}} \quad
	\subfloat[][]{\includegraphics[clip, width=0.25\columnwidth]{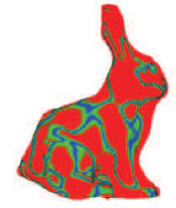}} \quad
	\subfloat[][]{\includegraphics[clip, width=0.25\columnwidth]{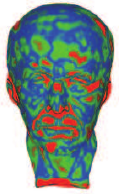}} \quad
	\subfloat[][]{\includegraphics[clip, width=0.25\columnwidth]{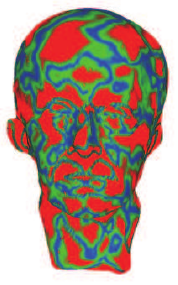}} \quad
	\subfloat[][]{\includegraphics[clip, width=0.25\columnwidth]{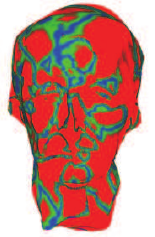}} \quad
	\subfloat[][]{\includegraphics[clip, width=0.25\columnwidth]{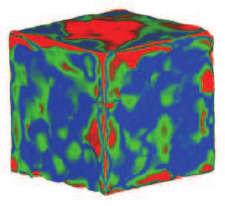}} \quad
	\subfloat[][]{\includegraphics[clip, width=0.25\columnwidth]{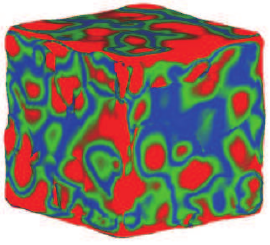}} \quad
	\subfloat[][]{\includegraphics[clip, width=0.25\columnwidth]{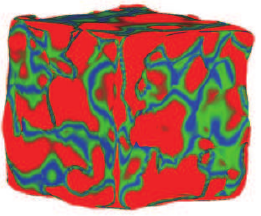}} 
\caption{Results of Experiment 2: (a--c), (d--f), and (g--i) are the isomorphic meshes after the fine local mapping of the rabbit, Max Planck, and the box models, respectively. From left to right, the noise parameters ($\delta$, $\sigma$) are (20, 1.0), (40, 3.0), and (60, 5.0)}
\label{fig:exp2_fine_mapping_result}
\end{center}
\end{figure}

\begin{figure}[tb]
\begin{center}
	\subfloat[][]{\includegraphics[clip, width=1.0\columnwidth]{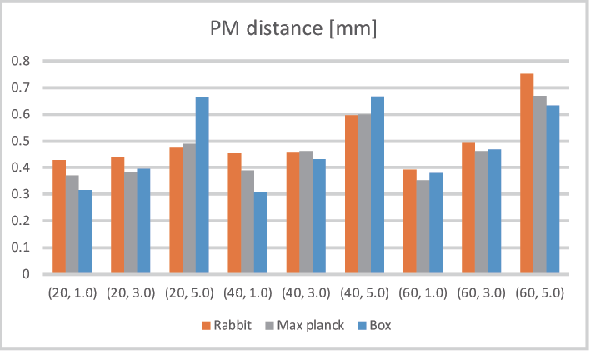}} 
\caption{Results of Experiment 2: PM distances of the isomorphic meshes after the fine local mapping of the rabbit, Max Planck, and the box models with noise in nine patterns ($\delta$, $\sigma$)}
\label{fig:exp2_pmdist}
\end{center}
\end{figure}

\subsection{Experiment 2: Isomorphic mesh generation using simulation data with noise}
To confirm the robustness of iMG to noise, we generated isomorphic meshes from the three simulation data containing artificial noise.
These simulation data were generated 
by adding uniformly distributed noise 
to simulation data without noise. 

The noise was controlled by two parameters $\delta$ and $\sigma$.
The parameter $\delta$ [$\%$]
The parameter $\sigma$ represented the range $[-\sigma, \sigma]$ of the random values.
By combining $\delta = \{20, 40, 60 \}$ and $\sigma = \{1.0, 3.0, 5.0 \}$, we generated nine patterns of the simulation data with noise.
Fig. \ref{fig:noised_input_points_for_simulation} shows the simulation data with noise for the three models when $(\delta, \sigma) = (20, 1.0), (40, 3.0),$ and $(60, 5.0)$.

Fig. \ref{fig:exp2_fine_mapping_result} shows the isomorphic meshes generated from these simulation data, and Fig. \ref{fig:exp2_pmdist} shows their PM distances using the final isomorphic meshes.
From these figures,
the shape-recovery accuracy decreased in inverse proportion to the increase in the ratio of added noise and noise variance.
However, all the models recovered the approximate shapes of the target objects.

\subsection{Experiment 3: Isomorphic mesh generation using measurement data}
\begin{figure}[tb]
\begin{center}
	\subfloat[][]{\includegraphics[clip, width=0.25\columnwidth]{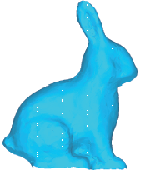}} \quad
	\subfloat[][]{\includegraphics[clip, width=0.2\columnwidth]{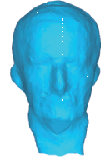}} \quad
	\subfloat[][]{\includegraphics[clip, width=0.25\columnwidth]{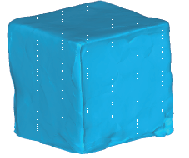}} 
\caption{Results of Experiment 3: Isomorphic meshes of (a) the rabbit, (b) Max Planck, and (c) the box models generated by iMG from the measurement data}
\label{fig:exp3_structured_mesh_model}
\end{center}
\end{figure}

\begin{figure}[tb]
\begin{center}
	\subfloat[][]{\includegraphics[clip, width=0.25\columnwidth]{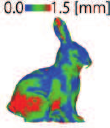}} \quad
	\subfloat[][]{\includegraphics[clip, width=0.2\columnwidth]{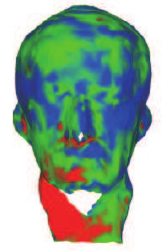}} \quad
	\subfloat[][]{\includegraphics[clip, width=0.25\columnwidth]{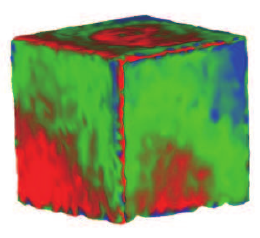}} \quad
	\subfloat[][]{\includegraphics[clip, width=0.25\columnwidth]{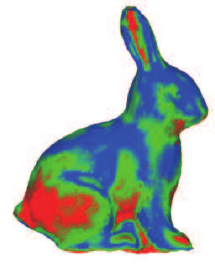}} \quad
	\subfloat[][]{\includegraphics[clip, width=0.2\columnwidth]{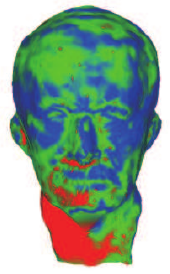}} \quad
	\subfloat[][]{\includegraphics[clip, width=0.25\columnwidth]{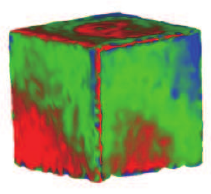}} 
\caption{Experiment 3: (a--c) Measurement surface meshes and (d--f) isomorphic meshes after fine local mapping of the rabbit, Max Planck, and the box models}
\label{fig:exp3_scanned_mesh_model_and_fine_mapping_result}
\end{center}
\end{figure}

The third experiment generated isomorphic meshes from the measurement data in Fig. \ref{fig:scanned_points}.
Figs. \ref{fig:exp3_structured_mesh_model} and \ref{fig:exp3_scanned_mesh_model_and_fine_mapping_result} (d)--(f) show the generated isomorphic meshes.
The colour map in Fig. \ref{fig:exp3_scanned_mesh_model_and_fine_mapping_result} was set such that the colour was close to red when the value of $d(\bm v, \bm G)$ was close to 1.5 [mm].
Table \ref{table:exp3_pmdist} presents the PM distances for each isomorphic mesh.

Additionally, iMG enabled the generation of the isomorphic meshes (Fig. \ref{fig:exp3_structured_mesh_model}) from an incomplete point cloud (Fig. \ref{fig:scanned_points}) while recovering the shapes of the missing parts.
Moreover, from Table \ref{table:exp3_pmdist}, the PM distance of the isomorphic mesh became smaller than that of the measurement mesh (Fig. \ref{fig:exp3_scanned_mesh_model_and_fine_mapping_result} (a)--(c)) for all the data.

\begin{table}[tb]
\centering
\caption{Results of Experiment 3: PM distances of the scanned surface meshes and isomorphic meshes after fine local mapping of the rabbit, Max Planck, and the box models}
\begin{tabular}{l|ll}
\hline
           & \multicolumn{2}{c}{PM distance [mm]} \\ \cline{2-3} 
            & Scanned mesh          & Isomorphic mesh                     \\ \hline
Rabbit     & 0.82                 & 0.77 $\pm$ 0.00         \\
Max Planck & 0.60                  & 0.57 $\pm$ 0.00          \\
Box        & 2.35                 & 1.47 $\pm$ 0.08         \\ \hline

\end{tabular}
\label{table:exp3_pmdist}
\end{table}

\subsection{Experiment 4: Comparison with AtlasNet using simulation data with or without noise}
\label{sec:comparison_with_atlasnet}
Unlike data-free iMG, AtlasNet is a data-driven method,
which generates an isomorphic mesh of a target object using a trained network in advance using the training dataset.
Therefore, if a class of objects is included in the training dataset, the shape of the target object is reconstructed with high accuracy.
Otherwise, the shape-recovery accuracy of the object is not guaranteed.
Owing to these characteristics of the data-driven method, the shape-recovery accuracy of AtlasNet is influenced by whether the class of an object to be modeled is included in the training data.
Therefore, a comparison was made for the two cases using the input point clouds of the trained and untrained classes.

When using AtlasNet, we used the pre-trained weights available on Github shown in \cite{groueix2018}.
To match the resolution of the spherical mesh used in AtlasNet, the number of vertices in the reference mesh of iMG was set to 2,562.

\subsubsection{Comparison between the two methods using input point clouds of trained classes}
As an input point cloud for this experiment, we selected a bench, chair, and car from the testing data used in \cite{groueix2018}.
Here, AtlasNet was applied to a point cloud that included 30,000 points, whereas the three simulation data (the rabbit, Max Planck, and the box) used in the experiment consisted of approximately 2,500 vertices.
Therefore, for the bench, chair, and car, AtlasNet was applied to the two types of input clouds with both 30,000 and 2,165 points.
Here, the bench, chair, and car were included in a bounding box with a length of 1 [mm], similar to the setting in the training data of AtlasNet.

\begin{figure*}[tb]
\begin{center}
\includegraphics[clip, width=2.0\columnwidth]{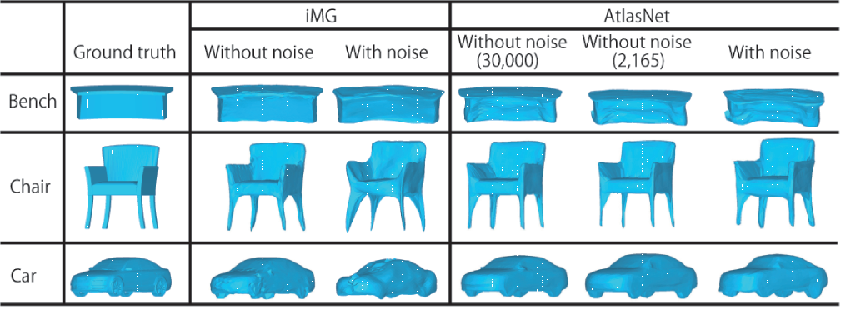}
\caption{Results of Experiment 4: Ground truths and isomorphic meshes generated using iMG and AtlasNet of the bench, chair, and car models. }
\label{fig:exp4_mapping_results_for_bench_shading}
\end{center}
\end{figure*}

\begin{figure}[tb]
\begin{flushleft}
	\quad
	\subfloat[][]{\includegraphics[clip, width=0.25\columnwidth]{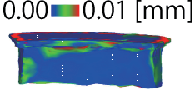}} \quad \quad
	\subfloat[][]{\includegraphics[clip, width=0.18\columnwidth]{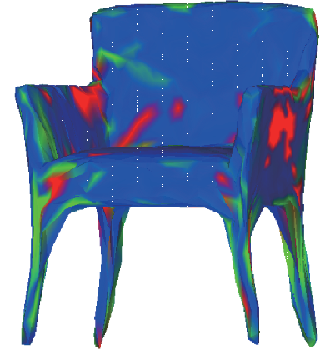}} \quad \quad
	\subfloat[][]{\includegraphics[clip, width=0.25\columnwidth]{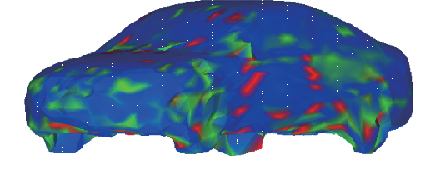}} \newline
	\quad
	\subfloat[][]{\includegraphics[clip, width=0.25\columnwidth]{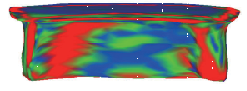}} \quad \quad
	\subfloat[][]{\includegraphics[clip, width=0.18\columnwidth]{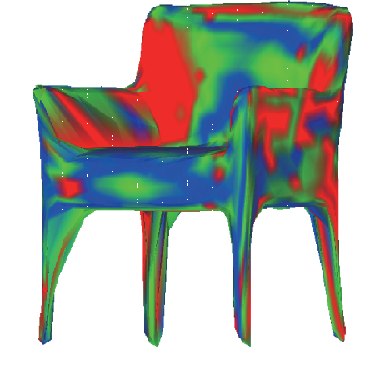}} \quad \quad
	\subfloat[][]{\includegraphics[clip, width=0.25\columnwidth]{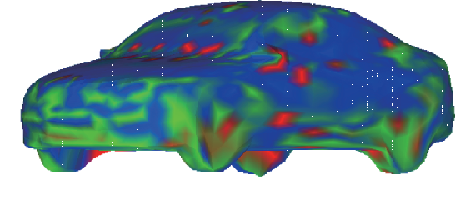}} \newline
	\quad
	\subfloat[][]{\includegraphics[clip, width=0.25\columnwidth]{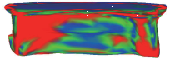}} \quad \quad
	\subfloat[][]{\includegraphics[clip, width=0.18\columnwidth]{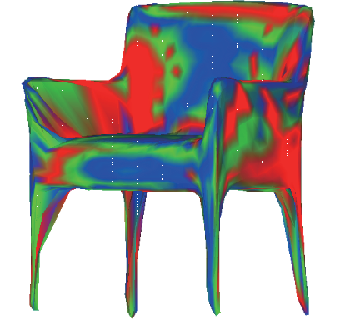}} \quad \quad
	\subfloat[][]{\includegraphics[clip, width=0.25\columnwidth]{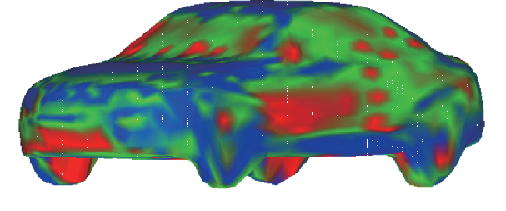}} 
	
\caption{Experiment 4: Isomorphic meshes generated by (a--c) iMG and (d--i) AtlasNet of the bench, chair, and car. 
The models shown in (d--f) and (g--i) are generated from input point clouds with 30,000 points and 2,165 points, respectively}
\label{fig:exp4_mapping_results_for_data_without_noise1}
\end{flushleft}
\end{figure}

Fig. \ref{fig:exp4_mapping_results_for_data_without_noise1} (a-c) shows the isomorphic meshes generated by iMG.
The isomorphic meshes generated by inputting the bench, chair, and car represented by 30,000 and 2,165 points to AtlasNet are shown in Fig. \ref{fig:exp4_mapping_results_for_data_without_noise1} (d--f) and (g--i), respectively.
Additionally, Fig. \ref{fig:exp4_mapping_results_for_bench_shading} shows the corresponding shaded models.
The colour maps of Fig. \ref{fig:exp4_mapping_results_for_data_without_noise1} were set such that the colour was close to red when the value of $d(\bm v, \bm G)$ was close to 0.01 [mm].
The third to eighth rows of Table \ref{table:exp4_pmdist} lists the PM distances.

Fig. \ref{fig:exp4_mapping_results_for_data_without_noise1} shows that there are more blue regions in the isomorphic meshes generated by iMG than those generated by AtlasNet. 
This indicates that the shape-recovery accuracy of the proposed iMG is higher than that of AtlasNet.
The third to eighth rows of Table \ref{table:exp4_pmdist} indicate that the PM distances of AtlasNet were equal to or lower for input point clouds with 30,000 points than for input point clouds with 2,165 points. 
Therefore, in AtlasNet, the shape-recovery accuracy was higher when the number of points in the input point cloud was aligned in advance with that in each point cloud of the training data, i.e., when the number of points was 30,000.
Even in this case,
the PM distances of iMG for the bench, chair, and car with 2,165 points
were lower than those of the AtlasNet with 30,000 points.

\begin{figure}[tb]
\begin{flushleft}
	\quad
	\subfloat[][]{\includegraphics[clip, width=0.25\columnwidth]{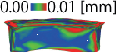}} \quad \quad
	\subfloat[][]{\includegraphics[clip, width=0.18\columnwidth]{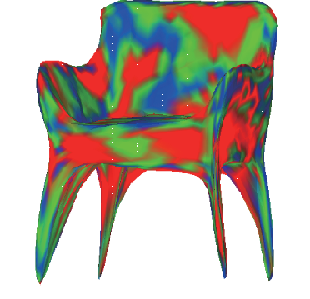}} \quad \quad
	\subfloat[][]{\includegraphics[clip, width=0.25\columnwidth]{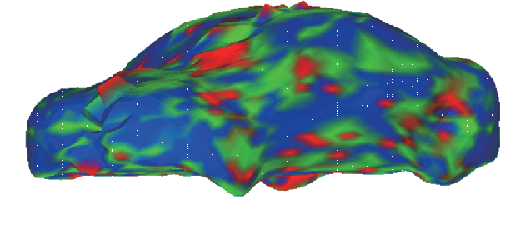}} \newline
	\quad
	\subfloat[][]{\includegraphics[clip, width=0.25\columnwidth]{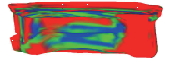}} \quad \quad
	\subfloat[][]{\includegraphics[clip, width=0.18\columnwidth]{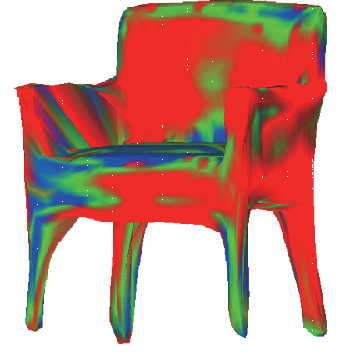}} \quad \quad
	\subfloat[][]{\includegraphics[clip, width=0.25\columnwidth]{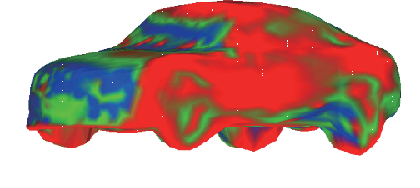}}

\caption{Results of Experiment 4: Isomorphic meshes generated using (a--c) iMG and (d--f) AtlasNet of the bench, chair, and car models with noise}
\label{fig:exp4_mapping_results_for_data_with_noise1}
\end{flushleft}
\end{figure}

To verify the robustness to noise, we added a uniform distribution noise of $\delta=40$ and $\sigma=0.03$ to the input point clouds with 2,165 points.
Figs. \ref{fig:exp4_mapping_results_for_data_with_noise1} (a-c) and (d-f) show the isomorphic meshes obtained by applying iMG and AtlasNet to the point clouds with noise. 
Fig. \ref{fig:exp4_mapping_results_for_bench_shading} shows the corresponding shaded models.
The colour map of Fig. \ref{fig:exp4_mapping_results_for_data_with_noise1} was set such that the colour was close to red when the value of $d(\bm v, \bm G)$ was close to 0.01 [mm].
The 12th to 14th rows of Table \ref{table:exp4_pmdist} lists the PM distances.
Fig. \ref{fig:exp4_mapping_results_for_data_with_noise1} shows that when noise was included, the red regions increased in the models obtained by iMG and AtlasNet. 
In particular, the shape-recovery accuracy of the edges was low in the result of iMG.
However, more blue regions existed in the result of iMG than in AtlasNet.
Table \ref{table:exp4_pmdist} shows that in the 12th to 14th rows, the PM distances were equal to or lower in iMG result than in the AtlasNet result.

\begin{figure*}[tb]
\begin{center}
\includegraphics[clip, width=1.5\columnwidth]{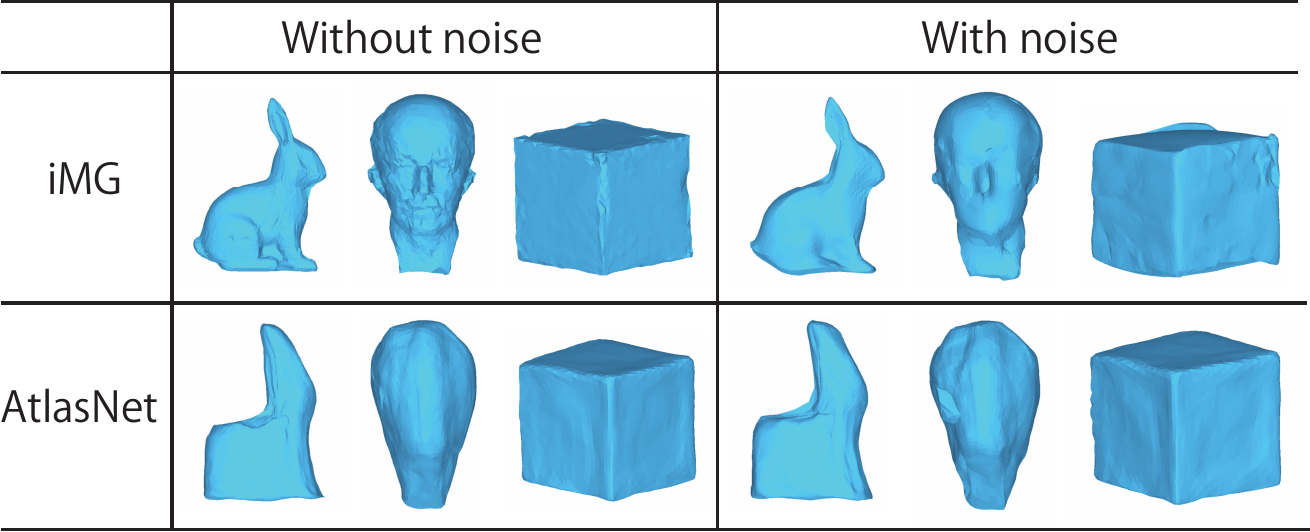}
\caption{Results of Experiment 4: Isomorphic meshes generated using iMG and AtlasNet of the rabbit, Max Planck, and the box models. }
\label{fig:exp4_mapping_results_for_box_rabbit_max_shading}
\end{center}
\end{figure*}

\begin{figure}[tb]
\begin{flushleft}
	\quad
	\subfloat[][]{\includegraphics[clip, width=0.25\columnwidth]{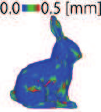}} \quad \quad
	\subfloat[][]{\includegraphics[clip, width=0.18\columnwidth]{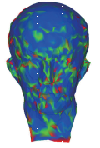}} \quad \quad
	\subfloat[][]{\includegraphics[clip, width=0.25\columnwidth]{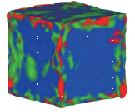}} \newline
	\quad
	\subfloat[][]{\includegraphics[clip, width=0.22\columnwidth]{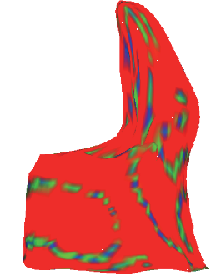}} \quad \quad \quad
	\subfloat[][]{\includegraphics[clip, width=0.18\columnwidth]{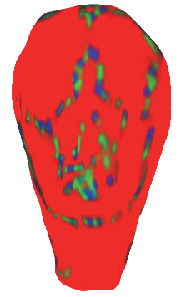}} \quad \quad
	\subfloat[][]{\includegraphics[clip, width=0.25\columnwidth]{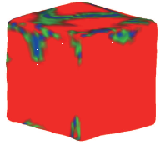}} 
	
\caption{Results of Experiment 4: Isomorphic meshes generated using (a--c) iMG and (d--f) AtlasNet of the rabbit, Max Planck, and the box models. }
\label{fig:exp4_mapping_results_for_data_without_noise2}
\end{flushleft}
\end{figure}

\begin{figure}[tb]
\begin{flushleft}
	\quad
	\subfloat[][]{\includegraphics[clip, width=0.2\columnwidth]{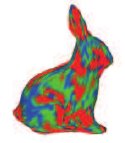}} \quad \quad
	\subfloat[][]{\includegraphics[clip, width=0.18\columnwidth]{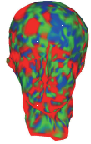}} \quad \quad
	\subfloat[][]{\includegraphics[clip, width=0.25\columnwidth]{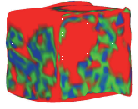}} \newline
	\quad
	\subfloat[][]{\includegraphics[clip, width=0.2\columnwidth]{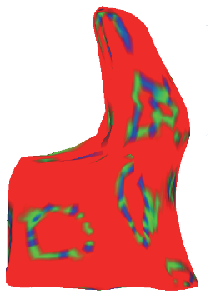}} \quad \quad
	\subfloat[][]{\includegraphics[clip, width=0.18\columnwidth]{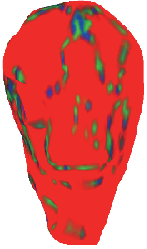}} \quad \quad
	\subfloat[][]{\includegraphics[clip, width=0.25\columnwidth]{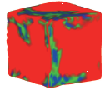}} 
	
\caption{Results of Experiment 4: Isomorphic meshes generated using (a--c) iMG and (d--f) AtlasNet of the rabbit, Max Planck, and the box models with noise}
\label{fig:exp4_mapping_results_for_data_with_noise2}
\end{flushleft}
\end{figure}

\begin{table}[tb]
\centering
\caption{Results of Experiment 4: PM distances of isomorphic meshes obtained using iMG and AtlasNet. The third to eleventh and 12th to 17th rows show the PM distances for input points without and with noise, respectively}
\begin{tabular}{cl|ll}
\hline
\multicolumn{1}{l}{}                                &                   & \multicolumn{2}{c}{PM distance [mm]} \\ \cline{3-4} 
\multicolumn{1}{l}{}                                &                   & iMG              & AtlasNet \\ \hline
\multicolumn{1}{c|}{\multirow{5}{*}{\begin{tabular}{c} Without\\ noise \end{tabular}}} & Bench (30,000)     & -                 & 0.010     \\
\multicolumn{1}{c|}{}                               & Bench (2,165)      & 0.006 $\pm$ 0.000   & 0.012     \\
\multicolumn{1}{c|}{} 										 & Chair (30,000)     & -                 & 0.010     \\
\multicolumn{1}{c|}{}                               & Chair (2,165)      & 0.008 $\pm$ 0.000   & 0.010     \\
\multicolumn{1}{c|}{} 										 & Car (30,000)     & -                 & 0.012     \\
\multicolumn{1}{c|}{}                               & Car (2,165)      & 0.011 $\pm$ 0.000   & 0.013     \\
\multicolumn{1}{c|}{}                               & Rabbit & 0.13 $\pm$ 0.00   & 1.70 \\
\multicolumn{1}{c|}{}                               & Max planck       & 0.16 $\pm$ 0.00   & 1.91    \\ 
\multicolumn{1}{c|}{}                               & Box               & 0.23 $\pm$ 0.01  & 1.74    \\ \hline
\multicolumn{1}{c|}{\multirow{4}{*}{\begin{tabular}{c} With\\ noise \end{tabular}}}    & Bench (2,165)     & 0.012 $\pm$ 0.001   & 0.014     \\
\multicolumn{1}{c|}{} 									   & Chair (2,165)     & 0.011 $\pm$ 0.000   & 0.012     \\
\multicolumn{1}{c|}{} 									   & Car (2,165)     & 0.013 $\pm$ 0.000   & 0.013     \\
\multicolumn{1}{c|}{}                               & Rabbit & 0.56 $\pm$ 0.02   & 2.17       \\
\multicolumn{1}{c|}{}                               & Max planck       & 0.58 $\pm$ 0.02   & 1.82    \\ 
\multicolumn{1}{c|}{}                               & Box               & 1.21 $\pm$ 0.06  & 1.28    \\ \hline
\end{tabular}
\label{table:exp4_pmdist}
\end{table}

\subsubsection{Comparison between the two methods using input point clouds of untrained classes}
Next, iMG was compared with AtlasNet via experiments using the three simulation data which belonged to untrained classes of AtlasNet.
Fig. \ref{fig:exp4_mapping_results_for_data_without_noise2} (a)--(c) and (d)--(f) show the isomorphic meshes obtained using iMG and AtlasNet from the input point clouds without noise.
Fig. \ref{fig:exp4_mapping_results_for_box_rabbit_max_shading} show the corresponding shaded models,
and the 9th to 11th rows of Table \ref{table:exp4_pmdist} lists their PM distances.
Here, before inputting the three simulation data to AtlasNet, we scaled down these data such that they could be included in a bounding box with a length of 1 [mm], similar to the setting in the training data of AtlasNet.
After scaling up the isomorphic meshes obtained using AtlasNet to their original scale, we calculated their shape-recovery accuracies.

As shown in Fig. \ref{fig:exp4_mapping_results_for_data_without_noise2} (a)--(c), iMG recovered the shape of the target object with high shape-recovery accuracy, whereas AtlasNet produced incomplete shapes of the objects of untrained classes (Fig. \ref{fig:exp4_mapping_results_for_data_without_noise2} (d)--(f)).
Fig. \ref{fig:exp4_mapping_results_for_data_without_noise2} (f) shows that AtlasNet could reconstruct the approximate shape of a box that was relatively close to the trained classes. 
However, even in this case, the shape-recovery accuracy of iMG was higher than that of AtlasNet.
Moreover, the ninth to eleventh rows of Table \ref{table:exp4_pmdist} show that iMG could represent the shape more accurately than AtlasNet.

Next, iMG and AtlasNet were applied to the simulation data with noise of $\delta=40$ and $\sigma=3.0$.
Fig. \ref{fig:exp4_mapping_results_for_data_with_noise2} (a)--(c) and (d)--(f) show the models obtained using iMG and AtlasNet,
and Fig. \ref{fig:exp4_mapping_results_for_box_rabbit_max_shading} shows the corresponding shaded models.
Moreover, their PM distances are shown in the 15th to 17th rows of Table \ref{table:exp4_pmdist}.
Fig. \ref{fig:exp4_mapping_results_for_data_with_noise2} (a)--(c) show that compared to Fig. \ref{fig:exp4_mapping_results_for_data_without_noise2} (a)--(c), the shape-recovery accuracy of iMG generally decreased.
However, 
as Table \ref{table:exp4_pmdist} shows in the 15th to 17th rows, the PM distances of iMG were shorter than those of AtlasNet.

%% file: discussion.tex
\section{Discussion}
Based on the experimental results in Section \ref{sec:experiment},
we discuss the shape-recovery accuracy of iMG and the robustness of iMG to noise in this section.

\subsection{Shape-recovery accuracy}
The results of Experiment 1
in Figs. \ref{fig:structured_mesh_models_for_simulation} and \ref{fig:mapping_results_for_simulation} and Table \ref{table:result_for_exp1}
% and Table \ref{table:result_for_exp1}
indicate that iMG enabled the generation of isomorphic meshes for input point clouds without noise, while maintaining the mesh structure of the reference mesh.
In particular, iMG achieved high accuracy in recovering the shape of curved surfaces with large curvatures.

%Generally, the shape-recovery accuracy for the target object with many edges often becomes lower than that of other objects with many curved surfaces.
%However, the characteristic of the isomorphic mesh for the box with many edges is different from that for other models.
%Therefore, in this section, we discuss about the results for the box with many edges.
By contrast,
as shown in the results using the box model,
the accuracy of recovering the edge shapes in the fine local mapping (Fig. \ref{fig:mapping_results_for_simulation} (i)) became worse than that of the coarse local mapping (Fig. \ref{fig:mapping_results_for_simulation} (h)).
In fine local mapping, the reference mesh was divided into local meshes based on the shape of the input point cloud.
In the box model, more local meshes were generated around the edges, as described in Section \ref{sec:surface_division}.
Therefore, we had to integrate more deformed local meshes into one mesh smoothly around the edges.

In local mesh integration, the vertex position in the overlapping region was determined by the weighted average of the positional coordinates, which were calculated using the mapping functions corresponding to the local meshes.
One mapping function used in iMG recovered one curved surface.
Here, the edge of the box model was a straight line formed by the intersection of two planes.
%When the mapping function is applied to the edge shape, the mapping function recovers an apploximate curved surface for the curved surfaces where the edge is.
%Therefore, it is not possible to accurately recover the edge shape by the mapping function. 
To accurately recover the edge shape, a local cloud should include points on one plane that the mapping function can recover.
In fine local mapping, as described in Section \ref{sec:surface_division}, the input point cloud was divided such that the shape of the local cloud was represented using a simple function of the polynomial surface. 
However, there was no guarantee that the points on the two planes were divided into different local clouds according to each plane.
Moreover, there was always an overlap between two neighbouring local clouds to connect the local meshes smoothly.
Therefore, around the edges, the points on the two planes were included in one local cloud.

In this case, if the number of points on one of the two planes was similar to that of the other, the MLP was optimized to approximate a mapping function for recovering a curved surface that approximated the edge shape.
Owing to this, the recovered edge shape became smoother than the edge shape of the ground truth.

Moreover, when there was a large bias in the number of points on each plane of the two planes, the MLP was optimized to approximate a mapping function to recover the shape of the plane with a larger number of points. 
In this case, because the shape of the local mesh deformed by the mapping function approached the plane, the edge shape included in the local mesh was not recovered.
Therefore,
a vertex on and around the edge
was difficult to obtain
by integrating the local meshes.
This difficulty resulted in the degradation of the accuracy of recovering the edge shape in the overlapping region.

\begin{figure}[tb]
\begin{center}
	\subfloat[][]{\includegraphics[clip, width=0.3\columnwidth]{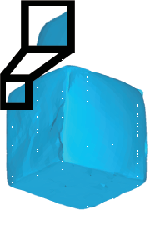}} \quad
	\subfloat[][]{\includegraphics[clip, width=0.35\columnwidth]{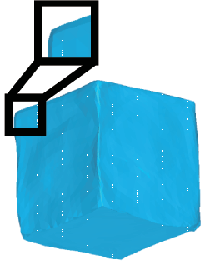}} 
\caption{Results of Experiment 4: Isomorphic meshes generated using (a) iMG and (b) AtlasNet from the input point cloud with artificial noise}
%\ecaption{Algorithm of the proposed method.}
\label{fig:box_iMG_AtlasNet}
\end{center}
\end{figure}

In Experiment 4, we compared the shape-recovery accuracy of iMG with that of AtlasNet.
Since the data-driven AtlasNet learned the shape characteristics of the edges in advance, the edge shape of the isomorphic mesh was represented if some noise was added to the edge of the input point cloud (Fig. \ref{fig:box_iMG_AtlasNet} (b)).
However, the data-free iMG generated isomorphic meshes only from the input point cloud. 
Therefore, when a small amount of noise was included in the input point cloud, recovering the edge shape became difficult.
As shown in Fig. \ref{fig:box_iMG_AtlasNet} (a), the edge shapes of the obtained isomorphic mesh of the box became smoother than the ground truth shape (Fig. \ref{fig:simulation_mesh} (c)).
From these results,
to improve the recovery accuracy of edge shapes, we considered combining data-driven and data-free frameworks in the future.

Moreover, in Experiments 1 and 4, the isomorphic meshes were generated using two types of reference meshes with 36,002 and 2,562 vertices.
Table \ref{table:result_for_exp1} and the 9th to 11th columns of Table \ref{table:exp4_pmdist} show the PM distances using the reference mesh with 36,002 and 2,562 vertices, respectively.
In comparison, the PM distances with 36,002 vertices were found to be 0.06-0.12 [mm] lower than those with 2,562 vertices.
This result indicated that
iMG improved the shape recovery
accuracy according to the increase in
the number of vertices in the reference mesh.

\subsection{Robustness to noise}
Experiments 2 and 3 were conducted to verify whether iMG can generate isomorphic meshes from two types of input point clouds with noise: that (Fig. \ref{fig:noised_input_points_for_simulation}) produced by adding artificial noise and that (Fig. \ref{fig:scanned_points}) obtained by scanning real models using a TrueDepth camera.
%Since these input clouds include noise, by the two experiments , we evaluate the robustness of generating the isomorphic meshes by iMG to the noise data.

The results of Experiment 2 in Figs. \ref{fig:exp2_fine_mapping_result} and \ref{fig:exp2_pmdist} show
%Fig. \ref{fig:exp2_fine_mapping_result}, 
%show that the shape-recovery accuracy of the isomorphic mesh created using the noisy input point cloud is generally lower than that of the noise-free input point cloud.
that the overall shape-recovery accuracy tended to decrease in inverse proportion to the increase in the ratio $\delta$ of the added noise and noise variance $\sigma$.
The maximum lengths of the bounding boxes covering the rabbit, Max Planck, and the box models were 98.0, 98.9, and 100 [mm], respectively.
Fig. \ref{fig:exp2_pmdist} indicates that 
when the maximum length is used as 
the object size, since all object sizes are almost 100 [mm], 
the PM distances for all objects are lower than 1.0$\%$ of the object size. 
This indicated that iMG was robust to noise.

Fig. \ref{fig:exp2_pmdist} also reveals that in the case of the box model, the PM distance changed depending on the value of $\sigma$ only when $\delta$ was fixed.
By contrast, when $\sigma$ is fixed, the PM distance is almost the same regardless of the values of $\delta$.
In the case of the rabbit and Max Planck models, 
the PM distance changed depending on the values of $\sigma$ and $\delta$.
Here, because the box model has only edges and planes, even when $\delta$ is increased, the effect of noise on the target object remains almost the same.
Alternatively, the rabbit and Max Plank models consisted of curved
surfaces of various shapes, and
if $\delta$ was increased, noise was added to these curved surfaces.
The PM distance of the two models increased.

In Experiment 3, 
the measurement mesh included measurement errors, such as noise and missing parts.
However, as shown in Table \ref{table:exp3_pmdist}, iMG enabled the generation of isomorphic meshes while reducing the influences of measurement errors.
However, the shape-recovery accuracy of the isomorphic mesh (in Fig. \ref{fig:exp3_scanned_mesh_model_and_fine_mapping_result} (d)--(f)) was almost the same as that of its corresponding measurement mesh (Fig. \ref{fig:exp3_scanned_mesh_model_and_fine_mapping_result} (a)--(c)). 
More specifically, some influence of the measurement errors remained in the isomorphic meshes.

\begin{figure}[tb]
\begin{center}
	\subfloat[][]{\includegraphics[clip, width=1.0\columnwidth]{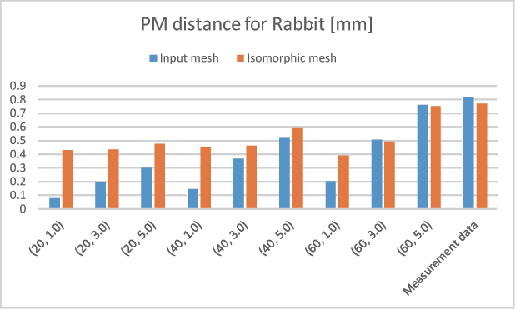}} \quad
	\subfloat[][]{\includegraphics[clip, width=1.0\columnwidth]{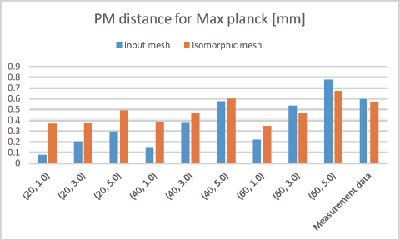}} \quad
	\subfloat[][]{\includegraphics[clip, width=1.0\columnwidth]{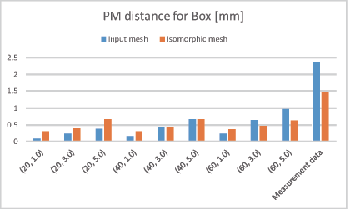}} 
\caption{PM distances for input meshes and isomorphic meshes obtained in Experiment 2 (nine patterns of ($\delta$, $\sigma$)) and Experiment 3 (measurement data)}
%\ecaption{Algorithm of the proposed method.}
\label{fig:PMdist_for_input_mesh}
\end{center}
\end{figure}

Moreover, we analysed the influence of noise and missing parts on the shape-recovery accuracy of iMG in detail. 
Fig. \ref{fig:PMdist_for_input_mesh} shows the PM distances between the isomorphic mesh obtained using iMG and its input mesh.
In Fig. \ref{fig:PMdist_for_input_mesh}, the input mesh was the surface mesh with noise corresponding to the input point cloud in Experiment 2 and a measurement mesh in Experiment 3.
When the noise was small, the PM distance of the isomorphic mesh was longer than that of the input mesh.
By contrast, when the noise was large ($\delta = 60$ and $\sigma > 3.0$), the PM distance of the isomorphic mesh was shorter than that of the input mesh in all models.

Since iMG was a data-free method, an approximation error always occurred when fitting curved surfaces into the input point cloud with and without noise.
When the noise was small, 
the approximation errors influenced shape-recovery accuracy.
Therefore, the shape-recovery accuracy of the isomorphic mesh was lower than that of the input mesh.
However, when the noise was large, the approximation by curved surfaces had the effect of reducing noise.
Since the effect of noise reduction was greater than that of the approximation error, the shape-recovery accuracy of the isomorphic mesh became higher than that of the input mesh when noise was large.

Moreover, from Fig. \ref {fig:PMdist_for_input_mesh}, the noise in the measurement mesh of the rabbit, Max Planck, and the box was close to $ (\delta, \sigma) = (60, 5.0) $, $(40, 5.0)$, and more than $(60, 5.0)$, respectively. 
Moreover, the measurement mesh of Experiment 3 included missing parts.
However, as shown in Fig. \ref{fig:exp3_structured_mesh_model}, iMG generated isomorphic meshes from the point cloud with noise while filling the missing parts.
Moreover,
the PM distances of the isomorphic meshes
were shorter than those of the measurement meshes
for all the models.
Therefore,
iMG was applicable for generating isomorphic meshes from the scanned point cloud using
mobile sensors.

\subsection{Comparing iMG and AtlasNet}
In Experiment 4, compared to AtlasNet, the proposed iMG generated isomorphic meshes with a short PM distance for the trained and untrained classes of AtlasNet.
Since AtlasNet was a data-driven method, the shape-recovery accuracy was high if the class of the input point cloud was included in the training data.
Owing to the characteristics of AtlasNet, in Experiment 4, AtlasNet achieved smooth approximations for the bench, chair, and car included in the training data.
Furthermore, although the class of the input point cloud was not trained, when the shape of the input point cloud was close to that of the trained class, shape-recovery accuracy was high.
For example, although the class of the box was not included in the training data, the shape of this class was similar to the cabinet included in the training data.
Therefore, the shape-recovery accuracy of the box was higher than that of the rabbit and Max Planck.
% and the box which is an untrained class but whose shape is close to the cabinet included in the training data.

However, the data-driven methods including AtlasNet were affected by the content of the training dataset. 
For example, as shown in Fig. \ref{fig:exp4_mapping_results_for_data_without_noise1} (d--f) and (g--i), if the number of points of each input point cloud used in the test was different from that used for training AtlasNet, their shape-recovery accuracy decreased.
Moreover, as shown in Fig. \ref{fig:exp4_mapping_results_for_data_without_noise2} (d--f), AtlasNet performed rough approximations for untrained classes.

However, because the data-free iMG estimated the shape only from the input point cloud, estimating various shapes with acceptable shape-recovery accuracy was possible.
Moreover, as shown in Table \ref{table:exp4_pmdist}, the shape-recovery accuracy of iMG was equal to or higher than that of AtlasNet for the objects in the trained class or objects with shapes close to that of the trained class. 

\subsection{Other isomorphic meshes generated using iMG}
\begin{figure}[tb]
\begin{center}
	\subfloat[][]{\includegraphics[clip, width=1.0\columnwidth]{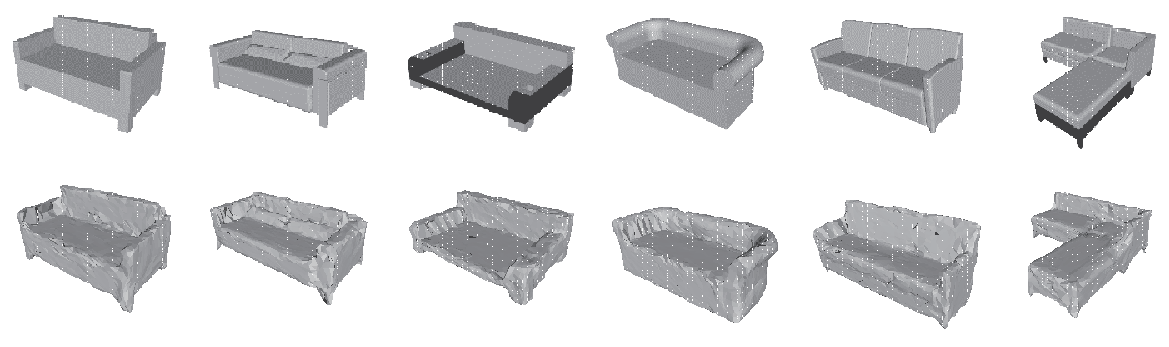}} \\
	\subfloat[][]{\includegraphics[clip, width=1.0\columnwidth]{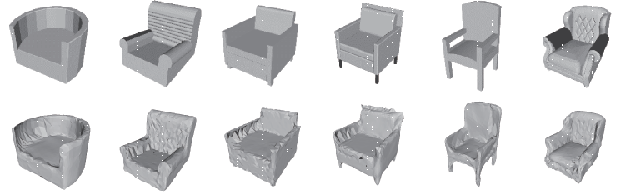}} \\
	\subfloat[][]{\includegraphics[clip, width=1.0\columnwidth]{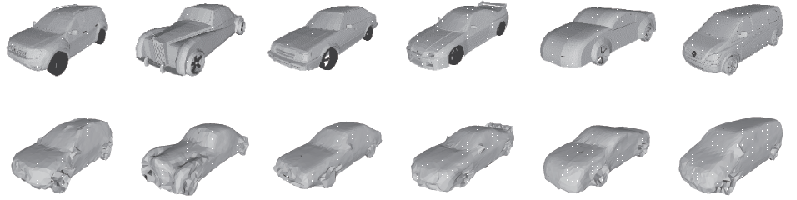}} \\
\caption{Lower models are the isomorphic meshes of (a) sofas, (b) chairs, and (b) cars, and upper models are their ground truths}
%\ecaption{Algorithm of the proposed method.}
\label{fig:other_results}
\end{center}
\end{figure}

To confirm the generalization of the proposed method, we generated isomorphic meshes of objects in multiple categories.
As the target categories, we selected sofas, chairs, and cars from ModelNet40 \cite{wu20153d}.
Fig. \ref{fig:other_results} shows the isomorphic meshes of the sofas, chairs, and cars generated using iMG.
In Fig. \ref{fig:other_results}, the models shown in the lower and upper rows are the generated isomorphic meshes and their ground-truths, respectively.
This figure confirmed that iMG generated isomorphic meshes of various objects without holes while recovering their ground-truth shapes.

\subsection{Limitation}
\label{sec:limitation}
Since iMG employed a spherical surface as the reference mesh, iMG always outputted isomorphic meshes with genus zero.
Therefore, the generation of isomorphic meshes using iMG was possible while filling the missing parts
when the input point cloud
included missing parts.
However, 
iMG also filled the holes that the objects originally had.
Moreover, in the missing parts where no points existed, the distribution of vertices in the isomorphic mesh became sparse.

When the number of points in a local cloud decreased,
overfitting to the local cloud occurred because only few points could be used for MLP training.
This overfitting often caused self-intersections of the edges in each local mesh.
Therefore, the isomorphic mesh generated from the local meshes after fine local mapping contained self-intersections.

%% file: conclusion.tex
\section{Conclusion}
We proposed iMG that generated an isomorphic mesh from a point cloud containing noise and missing parts.
In our iMG, a reference mesh was mapped to the input point cloud in a step-by-step manner. 
This process enabled flexible deformation, while maintaining the structure of the reference meshes.
Moreover, a data-free iMG required no training data in advance.

In the experiments, we generated isomorphic meshes using point clouds, with and without noise.
The experimental results showed that our iMG had a higher shape-recovery accuracy than that of AtlasNet.
Moreover, we generated isomorphic meshes using point clouds acquired by a mobile sensor.
From the experiments, we confirmed that iMG can output an isomorphic mesh with a shape close to the ground truth even if the input point cloud included noise and missing parts.

%% file: bare_jrnl_compsoc.bbl
\begin{thebibliography}{10}
\providecommand{\url}[1]{#1}
\csname url@samestyle\endcsname
\providecommand{\newblock}{\relax}
\providecommand{\bibinfo}[2]{#2}
\providecommand{\BIBentrySTDinterwordspacing}{\spaceskip=0pt\relax}
\providecommand{\BIBentryALTinterwordstretchfactor}{4}
\providecommand{\BIBentryALTinterwordspacing}{\spaceskip=\fontdimen2\font plus
\BIBentryALTinterwordstretchfactor\fontdimen3\font minus
  \fontdimen4\font\relax}
\providecommand{\BIBforeignlanguage}[2]{{%
\expandafter\ifx\csname l@#1\endcsname\relax
\typeout{** WARNING: IEEEtran.bst: No hyphenation pattern has been}%
\typeout{** loaded for the language `#1'. Using the pattern for}%
\typeout{** the default language instead.}%
\else
\language=\csname l@#1\endcsname
\fi
#2}}
\providecommand{\BIBdecl}{\relax}
\BIBdecl

\bibitem{liu2021pointguard}
H.~Liu, J.~Jia, and N.~Z. Gong, ``Pointguard: Provably robust 3d point cloud
  classification,'' in \emph{Proceedings of the IEEE/CVF Conference on Computer
  Vision and Pattern Recognition}, 2021, pp. 6186--6195.

\bibitem{wen2021airborne}
C.~Wen, X.~Li, X.~Yao, L.~Peng, and T.~Chi, ``Airborne lidar point cloud
  classification with global-local graph attention convolution neural
  network,'' \emph{ISPRS Journal of Photogrammetry and Remote Sensing}, vol.
  173, pp. 181--194, 2021.

\bibitem{liu2021fine}
X.~Liu, Z.~Han, Y.-S. Liu, and M.~Zwicker, ``Fine-grained 3d shape
  classification with hierarchical part-view attention,'' \emph{IEEE
  Transactions on Image Processing}, vol.~30, pp. 1744--1758, 2021.

\bibitem{mirbauer2021survey}
M.~Mirbauer, M.~Krabec, J.~Krivanek, and E.~Sikudova, ``Survey and evaluation
  of neural 3d shape classification approaches,'' \emph{IEEE Transactions on
  Pattern Analysis and Machine Intelligence}, 2021.

\bibitem{azcona2020interpretation}
E.~A. Azcona, P.~Besson, Y.~Wu, A.~Punjabi, A.~Martersteck, A.~Dravid, T.~B.
  Parrish, S.~K. Bandt, and A.~K. Katsaggelos, ``Interpretation of brain
  morphology in association to alzheimer’s disease dementia classification
  using graph convolutional networks on triangulated meshes,'' in
  \emph{International Workshop on Shape in Medical Imaging}.\hskip 1em plus
  0.5em minus 0.4em\relax Springer, 2020, pp. 95--107.

\bibitem{lakhili2019deformable}
Z.~Lakhili, A.~El~Alami, A.~Mesbah, A.~Berrahou, and H.~Qjidaa, ``Deformable 3d
  shape classification using 3d racah moments and deep neural networks,''
  \emph{Procedia computer science}, vol. 148, pp. 12--20, 2019.

\bibitem{george2022deep}
D.~George, X.~Xie, Y.~Lai, and G.~K. Tam, ``A deep learning driven active
  framework for segmentation of large 3d shape collections,''
  \emph{Computer-Aided Design}, vol. 144, p. 103179, 2022.

\bibitem{huang20213d}
H.~Huang, X.~Li, L.~Wang, and Y.~Fang, ``3d-metaconnet: Meta-learning for 3d
  shape classification and segmentation,'' in \emph{2021 International
  Conference on 3D Vision (3DV)}.\hskip 1em plus 0.5em minus 0.4em\relax IEEE,
  2021, pp. 982--991.

\bibitem{qiu2021dense}
S.~Qiu, S.~Anwar, and N.~Barnes, ``Dense-resolution network for point cloud
  classification and segmentation,'' in \emph{Proceedings of the IEEE/CVF
  Winter Conference on Applications of Computer Vision}, 2021, pp. 3813--3822.

\bibitem{wang2021learning}
X.~Wang, X.~Sun, X.~Cao, K.~Xu, and B.~Zhou, ``Learning fine-grained
  segmentation of 3d shapes without part labels,'' in \emph{Proceedings of the
  IEEE/CVF Conference on Computer Vision and Pattern Recognition}, 2021, pp.
  10\,276--10\,285.

\bibitem{xu2020weakly}
X.~Xu and G.~H. Lee, ``Weakly supervised semantic point cloud segmentation:
  Towards 10x fewer labels,'' in \emph{Proceedings of the IEEE/CVF conference
  on computer vision and pattern recognition}, 2020, pp. 13\,706--13\,715.

\bibitem{han2020occuseg}
L.~Han, T.~Zheng, L.~Xu, and L.~Fang, ``Occuseg: Occupancy-aware 3d instance
  segmentation,'' in \emph{Proceedings of the IEEE/CVF conference on computer
  vision and pattern recognition}, 2020, pp. 2940--2949.

\bibitem{yi2019gspn}
L.~Yi, W.~Zhao, H.~Wang, M.~Sung, and L.~J. Guibas, ``Gspn: Generative shape
  proposal network for 3d instance segmentation in point cloud,'' in
  \emph{Proceedings of the IEEE/CVF Conference on Computer Vision and Pattern
  Recognition}, 2019, pp. 3947--3956.

\bibitem{wang2019graph}
L.~Wang, Y.~Huang, Y.~Hou, S.~Zhang, and J.~Shan, ``Graph attention convolution
  for point cloud semantic segmentation,'' in \emph{Proceedings of the IEEE/CVF
  Conference on Computer Vision and Pattern Recognition}, 2019, pp.
  10\,296--10\,305.

\bibitem{yu2019partnet}
F.~Yu, K.~Liu, Y.~Zhang, C.~Zhu, and K.~Xu, ``Partnet: A recursive part
  decomposition network for fine-grained and hierarchical shape segmentation,''
  in \emph{Proceedings of the IEEE/CVF Conference on Computer Vision and
  Pattern Recognition}, 2019, pp. 9491--9500.

\bibitem{meng2019vv}
H.-Y. Meng, L.~Gao, Y.-K. Lai, and D.~Manocha, ``Vv-net: Voxel vae net with
  group convolutions for point cloud segmentation,'' in \emph{Proceedings of
  the IEEE/CVF international conference on computer vision}, 2019, pp.
  8500--8508.

\bibitem{jiang2018pointsift}
M.~Jiang, Y.~Wu, T.~Zhao, Z.~Zhao, and C.~Lu, ``Pointsift: A sift-like network
  module for 3d point cloud semantic segmentation,'' \emph{arXiv preprint
  arXiv:1807.00652}, 2018.

\bibitem{su2018splatnet}
H.~Su, V.~Jampani, D.~Sun, S.~Maji, E.~Kalogerakis, M.-H. Yang, and J.~Kautz,
  ``Splatnet: Sparse lattice networks for point cloud processing,'' in
  \emph{Proceedings of the IEEE conference on computer vision and pattern
  recognition}, 2018, pp. 2530--2539.

\bibitem{hanocka2019meshcnn}
R.~Hanocka, A.~Hertz, N.~Fish, R.~Giryes, S.~Fleishman, and D.~Cohen-Or,
  ``Meshcnn: a network with an edge,'' \emph{ACM Transactions on Graphics
  (TOG)}, vol.~38, no.~4, pp. 1--12, 2019.

\bibitem{chibane2020implicit}
J.~Chibane, T.~Alldieck, and G.~Pons-Moll, ``Implicit functions in feature
  space for 3d shape reconstruction and completion,'' in \emph{Proceedings of
  the IEEE/CVF Conference on Computer Vision and Pattern Recognition}, 2020,
  pp. 6970--6981.

\bibitem{genova2020local}
K.~Genova, F.~Cole, A.~Sud, A.~Sarna, and T.~Funkhouser, ``Local deep implicit
  functions for 3d shape,'' in \emph{Proceedings of the IEEE/CVF Conference on
  Computer Vision and Pattern Recognition}, 2020, pp. 4857--4866.

\bibitem{chabra2020deep}
R.~Chabra, J.~E. Lenssen, E.~Ilg, T.~Schmidt, J.~Straub, S.~Lovegrove, and
  R.~Newcombe, ``Deep local shapes: Learning local sdf priors for detailed 3d
  reconstruction,'' in \emph{European Conference on Computer Vision}.\hskip 1em
  plus 0.5em minus 0.4em\relax Springer, 2020, pp. 608--625.

\bibitem{chen2020bsp}
Z.~Chen, A.~Tagliasacchi, and H.~Zhang, ``Bsp-net: Generating compact meshes
  via binary space partitioning,'' in \emph{Proceedings of the IEEE/CVF
  Conference on Computer Vision and Pattern Recognition}, 2020, pp. 45--54.

\bibitem{gao2019sdm}
L.~Gao, J.~Yang, T.~Wu, Y.-J. Yuan, H.~Fu, Y.-K. Lai, and H.~Zhang, ``Sdm-net:
  Deep generative network for structured deformable mesh,'' \emph{ACM
  Transactions on Graphics (TOG)}, vol.~38, no.~6, pp. 1--15, 2019.

\bibitem{chen2019learning}
Z.~Chen and H.~Zhang, ``Learning implicit fields for generative shape
  modeling,'' in \emph{Proceedings of the IEEE/CVF Conference on Computer
  Vision and Pattern Recognition}, 2019, pp. 5939--5948.

\bibitem{qi2017pointnet}
C.~R. Qi, H.~Su, K.~Mo, and L.~J. Guibas, ``Pointnet: Deep learning on point
  sets for 3d classification and segmentation,'' in \emph{Proceedings of the
  IEEE conference on computer vision and pattern recognition}, 2017, pp.
  652--660.

\bibitem{qi2017pointnet++}
C.~R. Qi, L.~Yi, H.~Su, and L.~J. Guibas, ``Pointnet++: Deep hierarchical
  feature learning on point sets in a metric space,'' \emph{Advances in neural
  information processing systems}, vol.~30, 2017.

\bibitem{xie2021generative}
J.~Xie, Y.~Xu, Z.~Zheng, S.-C. Zhu, and Y.~N. Wu, ``Generative pointnet: Deep
  energy-based learning on unordered point sets for 3d generation,
  reconstruction and classification,'' in \emph{Proceedings of the IEEE/CVF
  Conference on Computer Vision and Pattern Recognition}, 2021, pp.
  14\,976--14\,985.

\bibitem{luo2021knn}
N.~Luo, Y.~Wang, Y.~Gao, Y.~Tian, Q.~Wang, and C.~Jing, ``knn-based feature
  learning network for semantic segmentation of point cloud data,''
  \emph{Pattern Recognition Letters}, vol. 152, pp. 365--371, 2021.

\bibitem{sheshappanavar2020novel}
S.~V. Sheshappanavar and C.~Kambhamettu, ``A novel local geometry capture in
  pointnet++ for 3d classification,'' in \emph{Proceedings of the IEEE/CVF
  Conference on Computer Vision and Pattern Recognition Workshops}, 2020, pp.
  262--263.

\bibitem{kazhdan2006poisson}
M.~Kazhdan, M.~Bolitho, and H.~Hoppe, ``Poisson surface reconstruction,'' in
  \emph{Proceedings of the fourth Eurographics symposium on Geometry
  processing}, vol.~7, 2006.

\bibitem{williams2021neural}
F.~Williams, M.~Trager, J.~Bruna, and D.~Zorin, ``Neural splines: Fitting 3d
  surfaces with infinitely-wide neural networks,'' in \emph{Proceedings of the
  IEEE/CVF Conference on Computer Vision and Pattern Recognition}, 2021, pp.
  9949--9958.

\bibitem{jiang2020local}
C.~Jiang, A.~Sud, A.~Makadia, J.~Huang, M.~Nie{\ss}ner, T.~Funkhouser
  \emph{et~al.}, ``Local implicit grid representations for 3d scenes,'' in
  \emph{Proceedings of the IEEE/CVF Conference on Computer Vision and Pattern
  Recognition}, 2020, pp. 6001--6010.

\bibitem{ummenhofer2021adaptive}
B.~Ummenhofer and V.~Koltun, ``Adaptive surface reconstruction with multiscale
  convolutional kernels,'' in \emph{Proceedings of the IEEE/CVF International
  Conference on Computer Vision}, 2021, pp. 5651--5660.

\bibitem{zhang2021learning}
J.~Zhang, Y.~Yao, and L.~Quan, ``Learning signed distance field for multi-view
  surface reconstruction,'' in \emph{Proceedings of the IEEE/CVF International
  Conference on Computer Vision}, 2021, pp. 6525--6534.

\bibitem{azinovic2021neural}
D.~Azinovi{\'c}, R.~Martin-Brualla, D.~B. Goldman, M.~Nie{\ss}ner, and
  J.~Thies, ``Neural rgb-d surface reconstruction,'' \emph{arXiv preprint
  arXiv:2104.04532}, 2021.

\bibitem{atzmon2020sal}
M.~Atzmon and Y.~Lipman, ``Sal: Sign agnostic learning of shapes from raw
  data,'' in \emph{Proceedings of the IEEE/CVF Conference on Computer Vision
  and Pattern Recognition}, 2020, pp. 2565--2574.

\bibitem{mi2020ssrnet}
Z.~Mi, Y.~Luo, and W.~Tao, ``Ssrnet: Scalable 3d surface reconstruction
  network,'' in \emph{Proceedings of the IEEE/CVF Conference on Computer Vision
  and Pattern Recognition}, 2020, pp. 970--979.

\bibitem{williams2019deep}
F.~Williams, T.~Schneider, C.~Silva, D.~Zorin, J.~Bruna, and D.~Panozzo, ``Deep
  geometric prior for surface reconstruction,'' in \emph{Proceedings of the
  IEEE/CVF Conference on Computer Vision and Pattern Recognition}, 2019, pp.
  10\,130--10\,139.

\bibitem{hanocka2020point2mesh}
R.~Hanocka, G.~Metzer, R.~Giryes, and D.~Cohen-Or, ``Point2mesh: a self-prior
  for deformable meshes,'' \emph{arXiv preprint arXiv:2005.11084}, 2020.

\bibitem{amenta1998new}
N.~Amenta, M.~Bern, and M.~Kamvysselis, ``A new voronoi-based surface
  reconstruction algorithm,'' in \emph{Proceedings of the 25th annual
  conference on Computer graphics and interactive techniques}, 1998, pp.
  415--421.

\bibitem{bernardini1999ball}
F.~Bernardini, J.~Mittleman, H.~Rushmeier, C.~Silva, and G.~Taubin, ``The
  ball-pivoting algorithm for surface reconstruction,'' \emph{IEEE transactions
  on visualization and computer graphics}, vol.~5, no.~4, pp. 349--359, 1999.

\bibitem{alexa2003computing}
M.~Alexa, J.~Behr, D.~Cohen-Or, S.~Fleishman, D.~Levin, and C.~T. Silva,
  ``Computing and rendering point set surfaces,'' \emph{IEEE Transactions on
  visualization and computer graphics}, vol.~9, no.~1, pp. 3--15, 2003.

\bibitem{ohtake2005multi}
Y.~Ohtake, A.~Belyaev, M.~Alexa, G.~Turk, and H.-P. Seidel, ``Multi-level
  partition of unity implicits,'' in \emph{Acm Siggraph 2005 Courses}, 2005,
  pp. 173--es.

\bibitem{kazhdan2013screened}
M.~Kazhdan and H.~Hoppe, ``Screened poisson surface reconstruction,'' \emph{ACM
  Transactions on Graphics (ToG)}, vol.~32, no.~3, pp. 1--13, 2013.

\bibitem{xiong2014robust}
S.~Xiong, J.~Zhang, J.~Zheng, J.~Cai, and L.~Liu, ``Robust surface
  reconstruction via dictionary learning,'' \emph{ACM Transactions on Graphics
  (TOG)}, vol.~33, no.~6, pp. 1--12, 2014.

\bibitem{moore2007survey}
P.~Moore and D.~Molloy, ``A survey of computer-based deformable models,'' in
  \emph{International Machine Vision and Image Processing Conference (IMVIP
  2007)}.\hskip 1em plus 0.5em minus 0.4em\relax IEEE, 2007, pp. 55--66.

\bibitem{miyauchi2018fast}
S.~Miyauchi, K.~Morooka, T.~Tsuji, Y.~Miyagi, T.~Fukuda, and R.~Kurazume,
  ``Fast modified self-organizing deformable model: Geometrical
  feature-preserving mapping of organ models onto target surfaces with various
  shapes and topologies,'' \emph{Computer methods and programs in biomedicine},
  vol. 157, pp. 237--250, 2018.

\bibitem{yu2018pu}
L.~Yu, X.~Li, C.-W. Fu, D.~Cohen-Or, and P.-A. Heng, ``Pu-net: Point cloud
  upsampling network,'' in \emph{Proceedings of the IEEE Conference on Computer
  Vision and Pattern Recognition}, 2018, pp. 2790--2799.

\bibitem{li2019lbs}
C.-L. Li, T.~Simon, J.~Saragih, B.~P{\'o}czos, and Y.~Sheikh, ``Lbs
  autoencoder: Self-supervised fitting of articulated meshes to point clouds,''
  in \emph{Proceedings of the IEEE/CVF Conference on Computer Vision and
  Pattern Recognition}, 2019, pp. 11\,967--11\,976.

\bibitem{groueix2018}
T.~Groueix, M.~Fisher, V.~G. Kim, B.~Russell, and M.~Aubry, ``{AtlasNet: A
  Papier-M\^ach\'e Approach to Learning 3D Surface Generation},'' in
  \emph{Proceedings IEEE Conf. on Computer Vision and Pattern Recognition
  (CVPR)}, 2018.

\bibitem{mescheder2019occupancy}
L.~Mescheder, M.~Oechsle, M.~Niemeyer, S.~Nowozin, and A.~Geiger, ``Occupancy
  networks: Learning 3d reconstruction in function space,'' in
  \emph{Proceedings of the IEEE/CVF Conference on Computer Vision and Pattern
  Recognition}, 2019, pp. 4460--4470.

\bibitem{peng2020convolutional}
S.~Peng, M.~Niemeyer, L.~Mescheder, M.~Pollefeys, and A.~Geiger,
  ``Convolutional occupancy networks,'' in \emph{European Conference on
  Computer Vision}.\hskip 1em plus 0.5em minus 0.4em\relax Springer, 2020, pp.
  523--540.

\bibitem{park2019deepsdf}
J.~J. Park, P.~Florence, J.~Straub, R.~Newcombe, and S.~Lovegrove, ``Deepsdf:
  Learning continuous signed distance functions for shape representation,'' in
  \emph{Proceedings of the IEEE/CVF Conference on Computer Vision and Pattern
  Recognition}, 2019, pp. 165--174.

\bibitem{yang2021deep}
M.~Yang, Y.~Wen, W.~Chen, Y.~Chen, and K.~Jia, ``Deep optimized priors for 3d
  shape modeling and reconstruction,'' in \emph{Proceedings of the IEEE/CVF
  Conference on Computer Vision and Pattern Recognition}, 2021, pp. 3269--3278.

\bibitem{niemeyer2019occupancy}
M.~Niemeyer, L.~Mescheder, M.~Oechsle, and A.~Geiger, ``Occupancy flow: 4d
  reconstruction by learning particle dynamics,'' in \emph{Proceedings of the
  IEEE/CVF international conference on computer vision}, 2019, pp. 5379--5389.

\bibitem{duan2020curriculum}
Y.~Duan, H.~Zhu, H.~Wang, L.~Yi, R.~Nevatia, and L.~J. Guibas, ``Curriculum
  deepsdf,'' in \emph{European Conference on Computer Vision}.\hskip 1em plus
  0.5em minus 0.4em\relax Springer, 2020, pp. 51--67.

\bibitem{yao20213d}
S.~Yao, F.~Yang, Y.~Cheng, and M.~G. Mozerov, ``3d shapes local geometry codes
  learning with sdf,'' in \emph{Proceedings of the IEEE/CVF International
  Conference on Computer Vision}, 2021, pp. 2110--2117.

\bibitem{cuturi2013sinkhorn}
M.~Cuturi, ``Sinkhorn distances: Lightspeed computation of optimal transport,''
  \emph{Advances in neural information processing systems}, vol.~26, pp.
  2292--2300, 2013.

\bibitem{wu20153d}
Z.~Wu, S.~Song, A.~Khosla, F.~Yu, L.~Zhang, X.~Tang, and J.~Xiao, ``3d
  shapenets: A deep representation for volumetric shapes,'' in
  \emph{Proceedings of the IEEE conference on computer vision and pattern
  recognition}, 2015, pp. 1912--1920.

\end{thebibliography}
